\documentclass[10pt,twocolumn,letterpaper]{article}

\usepackage{cvpr}              

\usepackage{url}
\usepackage{xcolor}         
\usepackage{graphicx}
\usepackage{colortbl}
\usepackage{pifont}
\usepackage{subcaption}
\usepackage{multirow}
\usepackage{amsmath}
\usepackage{booktabs}
\usepackage{dsfont}
\usepackage{rotating}
\usepackage{array}
\usepackage{makecell}
\usepackage{wrapfig}
\usepackage{enumitem}
\usepackage{amsthm}
\usepackage{utfsym} 
\usepackage{xspace} 
\usepackage{tcolorbox}
\usepackage{algorithm}
\usepackage{algorithmicx}
\usepackage{algpseudocode}
\usepackage{minibox}
\usepackage{mathtools}
\usepackage{multicol}

\usepackage{multibib}
\newcites{latex}{References}

\newtheorem{definition}{Definition}
\newtheorem{theorem}{Theorem}
\newcommand{\jh}[1]{{#1}}

\definecolor{iccvblue}{rgb}{0.21,0.49,0.74}
\definecolor{ourblue}{rgb}{0,0.5,1}
\definecolor{darkgreen}{rgb}{0.0,0.6,0.0}
\definecolor{beaublue}{rgb}{0.9, 0.95, 0.9}
\definecolor{blackish}{rgb}{0.2, 0.2, 0.2}
\definecolor{grayish}{rgb}{0.95, 0.95, 0.95}

\definecolor{LightCyan}{rgb}{0.88,1,0.88}
\definecolor{LightRed}{rgb}{1,0.88,0.88}

\definecolor{LightBlue}{rgb}{0.76274,0.92980,0.98431}

\definecolor{ourbluetwo}{rgb}{0.0,0.5,0.7}

\usepackage[bookmarks=false,pagebackref,breaklinks,colorlinks,allcolors=ourblue]{hyperref}

\newcommand{\red}[1]{{\color{red}#1}}

\newcommand{\negs}{$\!\!\!\!\!\!\!\!\!\!\!\!$}
\newcommand{\negsr}{$\!\!\!\!$}

\newcommand{\datas}[2]{$
\begin{subarray}{l}\displaystyle\text{#1}\\\text{\fontsize{9pt}{9pt}\selectfont#2}\end{subarray}$}

\makeatletter
\def\ps@myheadings{%
	\let\@oddfoot\@empty\let\@evenfoot\@empty
	\def\@evenhead{\thepage\hfil\slshape\leftmark}%
	\def\@oddhead{{\slshape\rightmark}\hfil\thepage}%
	\let\@mkboth\@gobbletwo
	\let\sectionmark\@gobble
	\let\subsectionmark\@gobble
}
\if@titlepage
\renewcommand\maketitle{\begin{titlepage}%
		\let\footnotesize\small
		\let\footnoterule\relax
		\let \footnote \thanks
		\null\vfil
		\vskip 60\p@
		\begin{center}%
			{\LARGE \@title \par}%
			\vskip 3em%
			{\large
				\lineskip .75em%
				\begin{tabular}[t]{c}%
					\@author
				\end{tabular}\par}%
			\vskip 1.5em%
			{\large \@date \par}
		\end{center}\par
		\@thanks\@notice
		\vfil\null
	\end{titlepage}%
	\setcounter{footnote}{0}%
}
\else
\renewcommand{\maketitle}{\par
	\begingroup
	\renewcommand\thefootnote{\@fnsymbol\c@footnote}%
	\def\@makefnmark{\rlap{\@textsuperscript{\normalfont\color{black}\@thefnmark}}}%
	\long\def\@makefntext##1{\parindent 1em\noindent
		\hb@xt@1.8em{%
			\hss\@textsuperscript{\normalfont\@thefnmark}}##1}%
	\if@twocolumn
	\ifnum \col@number=\@ne
	\@maketitle
	\else
	\twocolumn[\@maketitle]%
	\fi
	\else
	\newpage
	\global\@topnum\z@   
	\@maketitle
	\fi
	\thispagestyle{plain}\@thanks
	\endgroup
	\setcounter{footnote}{0}%
}
\makeatother

\usepackage{utfsym} 
\newcommand{\heart}{$\;\!$\usym{2665}}
\usepackage[misc]{ifsym}

\usepackage{xcolor}
\usetikzlibrary{calc}

\newcommand{\poincaretree}{%
\begin{tikzpicture}[scale=0.8, line cap=round, line join=round]

  \def\R{1.0}

  \foreach \k in {273,259,...,1}{
    \pgfmathsetmacro{\r}{\R*\k/260.0}
    \pgfmathsetmacro{\t}{\k/260.0}

    \pgfmathparse{
      ifthenelse(\t < 0.5,
        255 - 25*(\t/0.5),
        ifthenelse(\t < 0.75,
          230 - 35*((\t-0.5)/0.25),
          195 - 35*((\t-0.75)/0.25)
        )
      )
    }
    \let\red\pgfmathresult

    \pgfmathparse{
      ifthenelse(\t < 0.5,
        180 + 25*(\t/0.5),
        ifthenelse(\t < 0.75,
          205 + 20*((\t-0.5)/0.25),
          225 + 25*((\t-0.75)/0.25)
        )
      )
    }
    \let\green\pgfmathresult

    \pgfmathparse{
      ifthenelse(\t < 0.5,
        185 + 20*(\t/0.5),
        ifthenelse(\t < 0.75,
          205 - 25*((\t-0.5)/0.25),
          180 - 20*((\t-0.75)/0.25)
        )
      )
    }
    \let\blue\pgfmathresult

    \definecolor{tmpcol}{RGB}{\red,\green,\blue}
    \fill[tmpcol] (0,0) circle (1.03*\r);
  }

  \def\nsectors{6}
  \pgfmathsetmacro{\sectorstep}{360/\nsectors}

  \def\rtwo{0.965}
  \def\delta{12}
  \def\ratio{1.3}

  \pgfmathsetmacro{\a}{\ratio*\ratio - 1}
  \pgfmathsetmacro{\b}{-2*\ratio*\ratio*\rtwo*cos(\delta)}
  \pgfmathsetmacro{\cc}{\ratio*\ratio*\rtwo*\rtwo}
  \pgfmathsetmacro{\disc}{\b*\b - 4*\a*\cc}
  \pgfmathsetmacro{\rone}{(-\b - sqrt(\disc))/(2*\a)}

  \tikzset{
    rootnode/.style={
      circle,
      draw=black,
      fill=blue!35!gray,
      line width=0.6pt,
      inner sep=0pt,
      minimum size=4.0pt
    },
    treeedge/.style={black, line width=1.05pt}
  }

  \node[rootnode] (root) at (0,0) {};

  \foreach \s in {0,...,7}{
    \pgfmathsetmacro{\base}{\s*\sectorstep}

    \coordinate (l1-\s) at ({\base}:\rone);

    \coordinate (l2a-\s) at ({\base-\delta}:\rtwo);
    \coordinate (l2b-\s) at ({\base+\delta}:\rtwo);

    \ifnum\s=5
      \def\parentcol{orange!85!black!120}  
      \def\leafcol{red!45}
    \else\ifnum\s=6
      \def\parentcol{red!95!black}  
      \def\leafcol{red!80!black}
    \else\ifnum\s=7
      \def\parentcol{yellow!85!black} 
      \def\leafcol{yellow!65}
    \else
      \def\parentcol{blue!75}
      \def\leafcol{cyan!150!blue!135}
    \fi\fi\fi

    \draw[treeedge] (root) -- (l1-\s);
    \draw[treeedge] (l1-\s) -- (l2a-\s);
    \draw[treeedge] (l1-\s) -- (l2b-\s);

    \node[
      circle, draw=black, fill=\parentcol,
      line width=0.6pt, inner sep=0pt, minimum size=4.0pt
    ] at (l1-\s) {};

    \node[
      circle, draw=black, fill=\leafcol,
      line width=0.6pt, inner sep=0pt, minimum size=4.0pt
    ] at (l2a-\s) {};
    \node[
      circle, draw=black, fill=\leafcol,
      line width=0.6pt, inner sep=0pt, minimum size=4.0pt
    ] at (l2b-\s) {};
  }

\end{tikzpicture}%
}

\def\paperID{} 
\def\confName{CVPR}
\def\confYear{2026}

\title{
\hspace{0.2cm}
\begin{minipage}{0.04\textwidth}
\vspace{-0.35cm}
\poincaretree
\end{minipage}
\begin{minipage}{0.86\textwidth}\centering
\vspace{-0.35cm}
Hierarchically Robust Zero-shot Vision-language Models
\end{minipage}
\vspace{-0.75cm}
}

\author{%
Junhao Dong\textsuperscript{1,2}\quad Yifei Zhang\textsuperscript{3}\quad Hao Zhu\textsuperscript{4}\quad
Yew-Soon Ong\textsuperscript{1,2,\,\Letter} \quad
Piotr Koniusz\textsuperscript{5,4,\Letter}\\
\textsuperscript{1}Nanyang Technological University \quad
\textsuperscript{2}CFAR, IHPC, A*STAR\quad
\textsuperscript{3}Northwest Polytechnical University\\
\textsuperscript{4}Data61$\!${\color{red}\heart}CSIRO\quad
\textsuperscript{5}University of New South Wales\\
\vspace{-0.2cm}
{\tt\small \{junhao003, \!\!asysong\}@ntu.edu.sg, \{yifeiacc,\!\! allenhaozhu\}@gmail.com, piotr.koniusz@unsw.edu.au}
}

\begin{document}
\maketitle

\begingroup
\renewcommand\thefootnote{}
\footnotetext{ \hspace{-3ex} \Letter\ Corresponding authors.$\qquad$This paper is accepted by CVPR'26.}
\endgroup

\begin{abstract}
Vision-Language Models (VLMs) can perform zero-shot classification but are susceptible to adversarial attacks. While robust fine-tuning improves their robustness, existing approaches align fixed text embeddings with an image embedding,  sacrificing natural performance and robustness. A  robustness degradation also occurs when a model faces adversarial attacks targeting superclasses (parent classes, e.g., \texttt{mammal}) in addition to their base (leaf) classes (e.g., \texttt{cat}). Thus, to enhance adversarial robustness and leverage the inherent hierarchical properties of class space, we propose a novel adversarial fine-tuning framework based on hierarchical embeddings and several levels of adversarially robust alignment of image-text modalities. Additional mechanisms place visual embeddings at the desired depth of hierarchy, and we provide a theoretical connection between the depth of embedding in the hierarchy and the maximum viable margin size. Our model naturally realizes several margin sizes, boosting generalization of adversaries for robustification. As various trees with different parent labels can share the same leaf labels, we also consider aligning over multiple trees to boost semantic variety. Experiments across several datasets are performed.
\end{abstract}

\section{Introduction}

 Adversarial samples \cite{SzegedyZSBEGF13}\textemdash inputs nearly indistinguishable from clean images make Deep Neural Networks (DNNs) \cite{GoodfellowSS14, zhang2019adversarial}  and Vision-Language Models (VLMs) \cite{zhang2022towards, lu2023set,sun2026math} predict incorrect labels with high confidence. 
 VLMs are susceptible to adversarial threats due to their multi-modal nature, limiting their use in security-critical settings \cite{diaz2023connecting, fang2026disentangling}.

Recent studies on adversarial fine-tuning \cite{MaoGYWV23, wang2024pre, schlarmann2024robust} enhance zero-shot robustness in VLMs \cite{radford2021learning,kdd_adv,dong2025robust,simplex_adv} by a sample-wise alignment that matches adversarial features of each \textit{base class} (original category, \eg, \texttt{kit fox}) across image and text modalities,  neglecting auxiliary information contained by \textit{superclasses} (parent classes, \eg, \texttt{carnivore}$\leftarrow$\texttt{canine}$\leftarrow$\texttt{fox}). Thus, aligning an image embedding toward mere text embeddings of base classes compromises 
robustness because the adversarial samples generated in such a setting focus by design  on base classes only. Including more general superclasses to obtain adversarial samples leads to more general model robustification\footnote{The class hierarchy for ImageNet is publicly available\textemdash it follows WordNet \cite{miller1995wordnet} hierarchy. 
An arbitrary class hierarchy can also be built for a dataset's label space with a single ChatGPT prompt within seconds.}. 

\begin{figure}[!t]
	\vspace{-0.2cm}
	\begin{subfigure}[t]{0.49\linewidth} 
		\centering
		\includegraphics[width=1\linewidth]{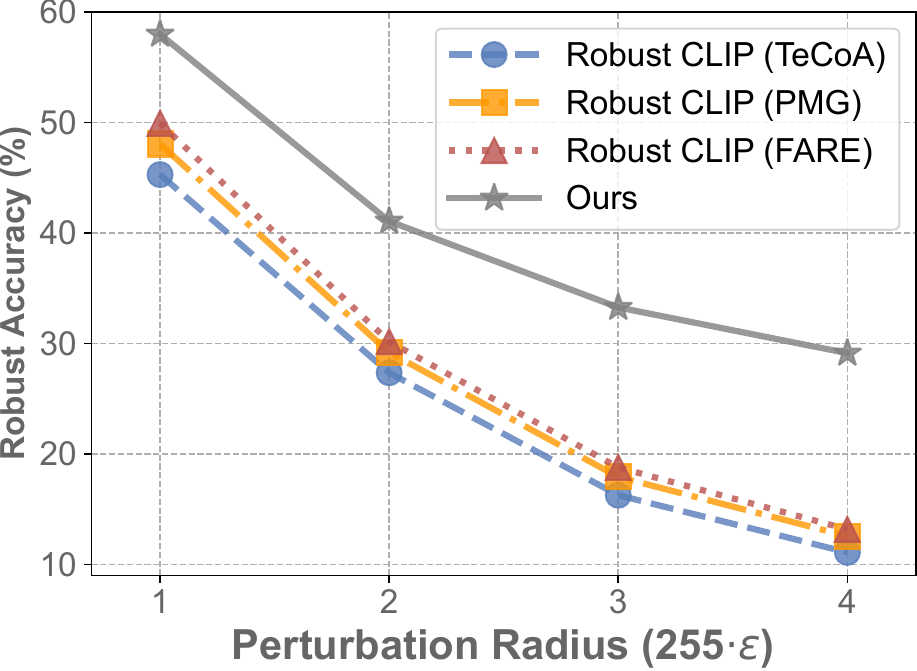}
		\vspace{-0.5cm}
		\caption{Superclass results}
		\label{fig:1_1}
		\vspace{0.2cm}
	\end{subfigure}
	\hfill
	\begin{subfigure}[t]{0.49\linewidth} 
		\centering
		\includegraphics[width=1\linewidth]{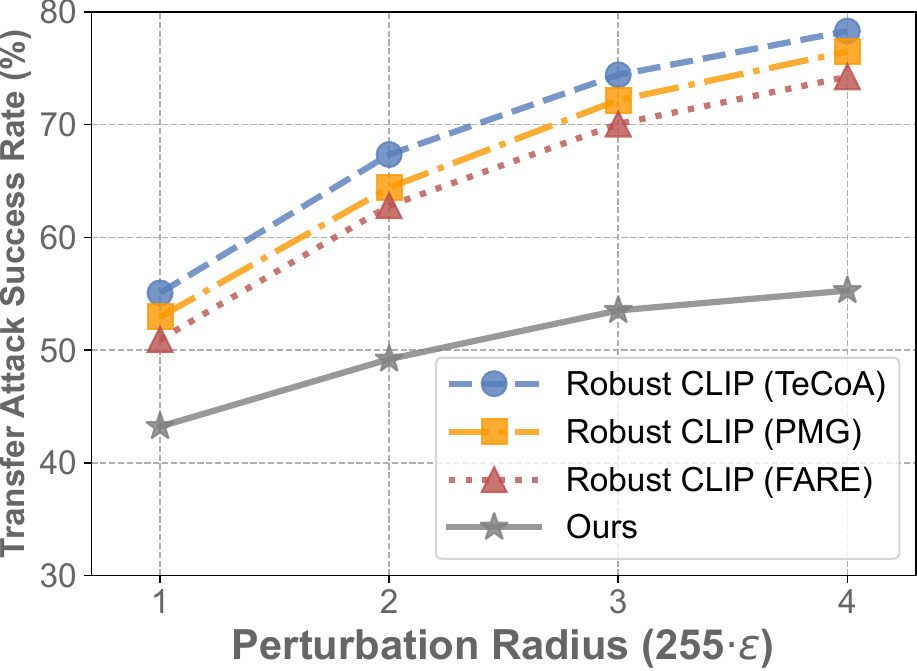}
		\vspace{-0.5cm}
		\caption{Base class results}
		\label{fig:1_2}
		\vspace{0.2cm}
	\end{subfigure}
	\vspace{-0.4cm}
	\caption{{\em Motivation (ImageNet \cite{deng2009imagenet} Val.)}  Fig. \ref{fig:1_1}: robust accuracy on superclasses (one level above the leaf classes). TeCoA \cite{MaoGYWV23}, PMG \cite{wang2024pre}, and FARE \cite{schlarmann2024robust} use adversaries derived from base classes and thus perform poorly on superclasses. Fig. \ref{fig:1_2}: transfer attack success rates on base classes. For our model, adversarial samples generated on superclasses were used. 
    In contrast, TeCoA \cite{MaoGYWV23}, PMG \cite{wang2024pre}, and FARE \cite{schlarmann2024robust}, the base classes are susceptible to attacks obtained from superclasses, highlighting the need for model robustification across class hierarchy (our model). 
    Examples of superclasses \&  leaves: (i) parent: \texttt {equine}, children: \texttt {horse, zebra}, (ii) parent: \texttt{car}, children: \texttt{freight car, taxi}, (iii) parent: \texttt{bag}, children: \texttt{backpack, purse}.
    }
	\label{fig:1}
	\vspace{-0.4cm}
\end{figure}

\begin{figure*}[!t]
\vspace{-0.2cm}
\begin{subfigure}[h]{0.52\linewidth} 
\includegraphics[height=3.5cm]{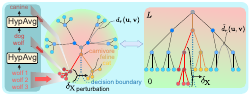}
\vspace{-0.5cm}
\caption{}
\label{fig:hypops}
\vspace{0.2cm}
\end{subfigure}
\hfill
	\begin{subfigure}[h]{0.16\linewidth} 
		\includegraphics[height=3.6cm]{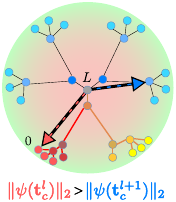}
		\vspace{-0.5cm}
		\caption{}
		\label{fig:norms0}
		\vspace{0.2cm}
	\end{subfigure}
	\hfill
	\begin{subfigure}[h]{0.285\linewidth} 
        \vspace{0.1cm}
		\includegraphics[height=3.6cm]{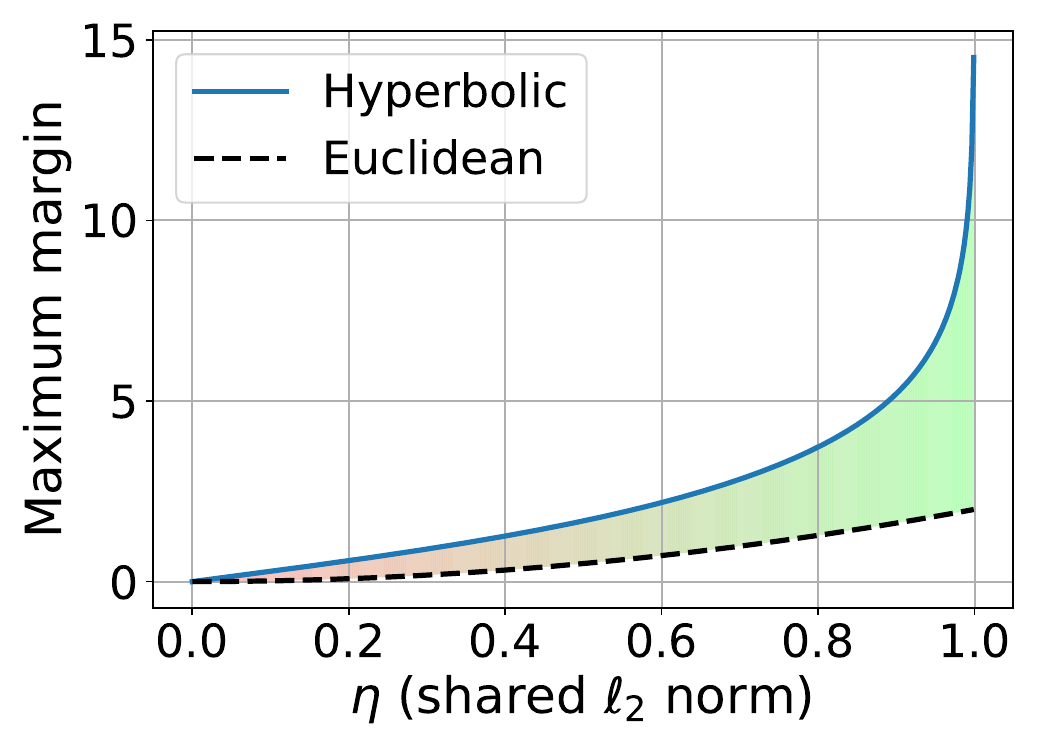}
        \vspace{-0.1cm}
		\vspace{-0.5cm}
		\caption{}
		\label{fig:hyp_margin}
		\vspace{0.2cm}
	\end{subfigure}
\vspace{-0.4cm}
\caption{{\em Understanding Hyperbolic Geometry.} Fig. \ref{fig:hypops}: Trees can be embedded in the Poincaré ball with low distortion ($\sim\!1\!+\!\epsilon$) \cite{sarkar2011low}. Thus, the distance between tree nodes $\tilde{d}_r(\mathbf{u},\mathbf{v})$ is the analogue of the Riemannian distance $d_r(\mathbf{u},\mathbf{v})$ between hyperbolic embeddings. The norm of the vector measured in the Poincaré ball is a proxy to the hierarchical level of a node in the tree (Fig. \ref{fig:norms0}) (we use ``reciprocal'': $L$ is the root level and $0$ is the leaves).
For the visual brach of CLIP, we have images with base class labels only. Thus, hyperbolic average (HypAvg) produces more generic embeddings  from the more specific embeddings (closer to the center than children), \eg, from  \texttt{wolf 1}, \texttt{wolf 2}, \texttt{wolf 3} it produces \texttt{wolf}, and from \texttt{wolf}, \texttt{dog} it produces \texttt{canine}. To produce a robust adversarial sample used for robustification, we compute $\delta_X$ (added to a sample) that makes nodes on the red path move across the decision boundary towards other categories (\eg, blue path).
Fig. \ref{fig:norms0}: The $\ell_2$ norm of  features (\eg, text $\psi(\cdot)$) controls the embedding level, \eg, {\color{red}$\lVert\psi(\mathbf{t}^l_c)\rVert_2$}$>${\color{ourblue}$\lVert\psi(\mathbf{t}^{l+1}_c)\rVert_2$} (red \vs blue arrow) has  order {\color{red}$0\!<\!l$}$<$ {\color{ourblue}$l\!+\!1\!<\!L$} in the Poincaré ball.
Fig. \ref{fig:hyp_margin}: Hyperbolic \vs Euclidean 
classifier's limit on viable margin size 
\wrt the $\ell_2$ norm of features ($\eta$). Our hyperbolic classifier can separate embeddings by an infinite margin (blue solid curve) when norm $\eta\!\rightarrow\!1/\sqrt{r}$ for curvature $r$. The Euclidean model (dashed black) has finite margin. Several  levels of hierarchy produce several margin sizes in $(0,\infty)$ to cover several levels of generalization, producing more general adversarial perturbation $\delta_X$ compared to the Euclidean model.
}
\vspace{-0.2cm}
\end{figure*}


To empirically support the above claim, for each adversarially fine-tuned CLIP model \cite{MaoGYWV23, wang2024pre, schlarmann2024robust}, we generate adversarial samples targeting the superclasses rather than the base classes of original samples, thereby creating more generic adversaries. 
Figure \ref{fig:1_1} shows that prior adversarial fine-tuning methods, which were trained to robustify base classes, have poor robustness classification on superclasses. Thus, adversarial learning on the base classes fails to generalize the robustness to superclasses. Figure \ref{fig:1_2} shows that adversarial samples generated at the superclass level can be effectively transferred to attack base classes, resulting in a considerable attack success rate, especially as the perturbation radius increases. This vulnerability to superclass adversaries shows that prior methods lose robustness due to the base classes usage only. We point out the benefit of achieving hierarchical robustness on image-text modalities by exploring the richness of hierarchical decision boundaries. 

To avoid the fixation on base categories, and achieve  hierarchically adversarially robust image-text modality alignment,  we propose a novel  framework based on hyperbolic embeddings \cite{hyperbolic_collapse} of image/category  hierarchies.

Figure \ref{fig:hypops} shows  we 
 encode children \& their parent labels across all levels of hierarchy. Text prompt embeddings of children labels are encouraged to lie closer to the  Poincaré boundary, whereas text prompt embeddings of parent labels are encouraged to lie closer to the Poincaré ball origin. 

Thus, embeddings of superclasses at level $l\!+\!1$ are aligned with the hyperbolic average of embeddings of images of a given class at level $l$ in the mini-batch ($l\!=\!0$: leaves, $L$: root). The hyperbolic average has the known property that from two/more children vectors, it yields a lower $\ell_2$ norm compared to children's norms. Thus, the hyperbolic average produces more general concepts (at level $l\!+\!1$ closer to the root) from more specific concepts (at level $l$) as the value of the norm controls the node depth (Fig. \ref{fig:norms0}).

Moreover, Figure \ref{fig:hyp_margin} shows that the hyperbolic classifier enjoys
 the limit on viable margin size
 growing exponentially (unlike the Euclidean classifier) to infinity (larger separation) with norm $\eta$ (depth) growing toward tree leaves. Thus, we employ several hierarchical levels of classifier to reformulate traditional sample-wise adversarial fine-tuning into its hierarchical counterpart, capturing rich adversarial information from hierarchical decision boundaries across several levels of generalization due to varying margin sizes.

Our core contributions  are listed below:
%
%


\renewcommand{\labelenumi}{\roman{enumi}.}
\begin{enumerate}[leftmargin=0.4cm]
\item  Unlike standard adversarial fine-tuning that aligns image and text embeddings under flat class structure, we cast adversarial fine-tuning as a hierarchical scheme that exploits rich hierarchical decision boundaries from child and parent categories that realize several varying margin sizes due to our hyperbolic design. This leads to more general adversarial perturbations for robustification compared to the Euclidean model. We also demonstrate theoretically  what is 
the limit on viable margin size.  

\item 
We analyze adversaries against superclass classification across varying perturbation radii. Our observations reveal that as the adversarial learning fixates on the base classes, it fails to generalize robustness to superclasses. Thus, we propose to leverage the hyperbolic embedding to exploit class hierarchies and image hierarchies. 
We propose mechanisms to deploy image and text embeddings meaningfully across several hierarchical levels.

\item We show on 15 datasets that our method  outperforms previous methods. 
 We also explore adversarially robust image-text retrieval,  captioning, and medical tasks. 
\end{enumerate}

\section{Background}

\noindent\textbf{Related works.}
Adversarial training enhances the model's robustness against attacks by integrating adversaries into training \cite{MadryMSTV18, zhang2019theoretically, Dong_2023_CVPR, dong2024adversarially, dong2024adversariallydistill, 10177878, dong2024robust, dong2024survey}. As adversarial training of large VLMs is costly \cite{radford2021learning}, adversarial fine-tuning \cite{MaoGYWV23, schlarmann2024robust, dong2026allies, dongrobust, dong2025confound, dongtug} is a better alternative. Mao \etal \cite{MaoGYWV23} employed text-guided contrastive learning for adversarial image-text embedding matching. Wang \etal \cite{wang2024pre} included feature-level regularization. Schlarmann \etal \cite{schlarmann2024robust} developed an unsupervised robust learning framework for downstream tasks. Li \etal \cite{li2024one} used one prompt word to boost robustness. However, these existing methods 
overlook the benefit of more generalized adversaries from hierarchical decision boundaries. 
See Appendix \ref{supp:Extend_RW} for more related works.


\vspace{0.1cm}
\noindent\textbf{Adversarial fine-tuning of VLM.} 
CLIP \cite{radford2021learning} uses image and text encoders parameterized by $\boldsymbol{\theta}_{\text{X}}$ and $\boldsymbol{\theta}_{\text{T}}$. 
It aligns visual and text embeddings in a shared feature space. The image and text encoders, $f\!:\! \mathcal{X}\!\rightarrow\!\mathbb{R}^{d}$ and $g\!:\!\mathcal{T}\!\!\rightarrow\!\mathbb{R}^{d}$, map input images $\mathbf{x}\!\in\!\mathcal{X}$ and input texts $\mathbf{t}\!\in\!\mathcal{T}$ to $d$-dimensional feature embeddings. 
By encoding image-text pairs $(\mathbf{x}, \mathbf{t})$ into a shared feature space, CLIP facilitates alignment between the two modalities. 
The probability ${p}_{c}(\mathbf{x})$ that an image $\mathbf{x}$ belongs to the base category of $c\!\in\!\{1,\ldots,C\}$ is defined as:
\begin{equation}
	\begin{aligned}
		{p}_{c}(\mathbf{x}) =  \frac{\operatorname{exp}\big[\operatorname{cos}\big(f(\mathbf{x}), g(\mathbf{t}_c)\big)\,\big]}{\sum_{c'=1}^{C}\operatorname{exp}\big[\operatorname{cos}\big(f(\mathbf{x}), g(\mathbf{t}_{c'})\big)\,\big]},
		\label{eq:1}
	\end{aligned}
\end{equation}
where $\operatorname{exp(\cdot)}$ and $\operatorname{cos(\cdot)}$ are the exponential function and the cosine similarity.  Tokenizer $h(\cdot)$ produces tokenized text embedding, $\mathbf{t}_c\!\!=\!h(\text{``}\texttt{[Context]} \texttt{[CLASS}_c\texttt{]}\text{''})$, \eg, ``\texttt{This is a photo of a} $[\texttt{CLASS}_c]$'' which is passed via the text encoder $g(\cdot)$ to produce a text embedding serving as a text reference for class $c$. Let $\mathbf{p} = [{p}_{1},\ldots, {p}_{C}]^\top\!\!\in\![0, 1]^{C}\!$. To enhance zero-shot robustness, standard adversarial fine-tuning (TeCoA \cite{MaoGYWV23}) can be applied to CLIP's image-text pairs $( \mathbf{x}, c)$ from the dataset $\mathcal{D}$ as follows:
\begin{equation}
	\begin{aligned}
		\min\limits_{\boldsymbol{\theta}_{\text{X}}} \,\mathbb{E}_{\left( \mathbf{x}, c\right)\sim \mathcal{D} }\left[ \max\limits_{\left\| \boldsymbol{\delta}_{\text{X}} \right\|_{\infty}\leq\epsilon_{\text{X}}}\mathcal{L}_{\text{CE}}\big( \mathbf{p}(\mathbf{x} + \boldsymbol{\delta}_{\text{X}}), \mathbf{y}(c) \big)  \right],
		\label{eq:2}
	\end{aligned}
\end{equation}
where the perturbation $\boldsymbol{\delta}_{\text{X}}$ is added to the image $\mathbf{x}$ to produce the adversarial sample $\mathbf{\hat{x}}\!\!=\!\!\mathbf{x}\!+\!\boldsymbol{\delta}_{\text{X}}$ within an $\ell_{\infty}$-norm ball of radius $\epsilon_{\text{X}}$. The target class $c$ is represented by a one-hot vector $\mathbf{y}(c) \!=\! [\mathds{1}(c=1),\ldots,\mathds{1}(c=C)]^\top\!\!\in\!\{0,1\}^C$. The inner maximization generates adversaries while the outer minimization reduces the empirical risk over adversaries. The adversary generation consists of perturbing visual inputs by maximizing the Cross-Entropy (CE) loss $\mathcal{L}_\text{CE}$ via iterative Projected Gradient Descent (PGD) \cite{MadryMSTV18}:
\begin{equation}
	\small
	\begin{aligned}
		\!\!\!\mathbf{\hat{x}}^{(i+1)} \!\!=\! \Pi_{\mathbb{B}(\mathbf{x}, \epsilon_{\text{X}})} \!\bigg[  \mathbf{\hat{x}}^{(i)} \!\!+\! \alpha_{\text{X}} \!\cdot\! \operatorname{sgn} \!\Big[ \nabla_{\mathbf{\hat{x}}^{(i)}}\mathcal{L}_{\text{CE}}\big( \mathbf{p}(\mathbf{\hat{x}}^{(i)}), \mathbf{y}(c) \big)\Big]  \bigg],\!\!
		\label{eq:3}
	\end{aligned}
\end{equation}
where $\operatorname{sgn}(\cdot)$ is the sign function, and $\alpha_{\text{X}}$ is the PGD step size. The projection operator $\Pi_{\mathbb{B}(\mathbf{x}, \epsilon_{\text{X}})}$ constraints the perturbed input to the $\ell_{\infty}$-norm hyperball of radius $\epsilon_{\text{X}}$ centered at $\mathbf{x}$. Adversarial initialization uses random perturbation: $\mathbf{\hat{x}}^{(0)} \sim \mathbf{x} + 0.001\cdot\mathcal{N}(\mathbf{0},\mathbf{I})$. The final adversarial sample $\mathbf{\hat{x}}=\mathbf{\hat{x}}^{(m)}$ is obtained after $m$ iterations.

\vspace{0.1cm}
\noindent\textbf{Poincaré ball model.}
\textit{Hyperbolic space} is a complete and connected Riemannian manifold with constant negative sectional curvature \cite{cannon1997hyperbolic}. We employ the Poincaré ball model $\mathbb{D}_r^d\!:=\!\left\{\boldsymbol{\phi} \in \mathbb{R}^d \mid\|\boldsymbol{\phi}\|_2^2<\frac{1}{r}\right\}$, 
where $\frac{1}{r}\!>\!0$ is the radius of the ball, and $\|\cdot\|_2$ is the $\ell_{2}$-norm. The Poincaré ball is endowed with a Riemannian metric  $g_{\mathbb{D}}(\boldsymbol{\phi})\!=\!\frac{4r}{\left(1-\|\boldsymbol{\phi}\|^2\right)^2}\, g_{\mathbb{E}}$, where $\boldsymbol{\phi}\!\in\! \mathbb{D}_r^d$ and $g_{\mathbb{E}}$ is the canonical Euclidean metric. 

\begin{figure}[!t]
\centering
\includegraphics[width=1\linewidth]{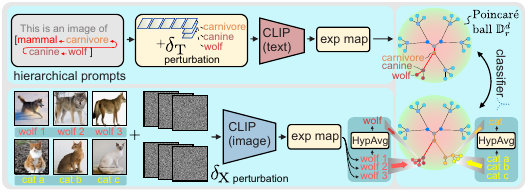}
\vspace{-0.6cm}
\caption{{\em Our pipeline.} Text  \& image CLIP encoders are used. For text,  we look up each class in a hierarchical tree (\eg, WorldNet for ImageNet) and extract the path from the leaf category to the root, \eg, animal$\leftarrow$mammal$\leftarrow$canine$\leftarrow$wolf. For each category level, we form one text prompt and encode it with CLIP.  The exponential map elevates embeddings from the Euclidean space into the hyperbolic space (Poincaré ball). HypAvg is hyperbolic averaging that generalizes specific embeddings (\eg, {\em wolf 1, wolf 2, wolf 3}) into a more generic parent embedding of {\em wolf}. Using the Riemannian distance, the text and visual ``wolf'' paths are aligned by a classifier while pushed away from other paths. Adversarial perturbations $\boldsymbol{\delta}_T$ and $\boldsymbol{\delta}_X$ are obtained on such a hierarchical classifier by the PGD \cite{MadryMSTV18}  for the robust model fine-tuning.}
\label{fig:pipe}
\vspace{-0.3cm}
\end{figure}

\section{Proposed Method}
Below, we introduce our adversarial fine-tuning for zero-shot hierarchically adversarially robust VLMs. 
Figure \ref{fig:pipe} shows our pipeline. Appendix \ref{supp:pipe} contains additional details. 

\vspace{0.1cm}
\noindent\textbf{Hyperbolic Embedding.} 
We represent the hierarchical structure inherent to each image-text pair by the Poincaré ball model to embed class hierarchy. 
Levels $1\!\leq\! l\!\leq\! L$ indicate superclasses while $l\!=\!0$ are base classes for  the total depth $L$. 
The tokenized text of the base class or superclass $c_l$ is  $ \mathbf{t}_c^l\! = \!h(\text{``}\texttt{[Context]} \texttt{[SUPERCLASS}^l_c\texttt{]}\text{''})$ where $h(\cdot)$ is a tokenizer. There exist $C_l$ classes at level $l$. 
For ImageNet, we use the WordNet \cite{miller1995wordnet} class hierarchy by extracting hypernyms of synsets as superclasses. In the absence of predefined hierarchical taxonomies in other datasets, we generate superclasses via  ChatGPT-4o \cite{openai_chatgpt_2024}. We prompt  LLM (takes few seconds) to get contextually appropriate superclasses for each base class, followed by manual reviews to ensure semantic quality. See Appendix \ref{supp:Superclass} for details.

To facilitate operations within hyperbolic space, we establish a bi-directional mapping between the Euclidean space $\mathbb{R}^d$ (CLIP embeddings) and the hyperbolic space $\mathbb{D}^d_r$. The exponential map in Definition \ref{def:Exp_Log_Maps} embeds Euclidean vectors into hyperbolic space (the log map is the inverse).

\vspace{-0.1cm}
\begin{definition}[\textbf{Exponential \& Logarithmic Maps}]
\label{def:Exp_Log_Maps}

The exponential map $\operatorname{exp}^{r}_{\mathbf{u}}(\cdot)$ is a function from $T_{\mathbf{u}} \mathbb{D}_r^d \!\cong\! \mathbb{R}^d$ to $\mathbb{D}_c^d$:
\vspace{-0.1cm}
\begin{equation}
	\begin{aligned}
    \exp _{\mathbf{u}}^r(\mathbf{v})=\mathbf{u} \oplus\Big[\tanh \Big(\frac{\sqrt{r}\lambda_{\mathbf{u}}^r}{2}\|\mathbf{v}\|_2\Big) \frac{\mathbf{v}}{\sqrt{r}\|\mathbf{v}\|_2}\Big],
    \label{eq:exp_map}
    \end{aligned}
\end{equation}
where $\oplus$ is the (differentiable) Möbius addition, and $\lambda_{\mathbf{u}}^r$ denotes the conformal factor for distance scaling. The (inverse) logarithmic map $\operatorname{log}^{r}_{\mathbf{u}}(\cdot)$ establishes an inverse projection from $\mathbb{D}_c^d$ to $T_{\mathbf{u}} \mathbb{D}_r^d \cong \mathbb{R}^d$, which is defined as:
\vspace{-0.2cm}
\begin{equation}
    \!\!\!\!\!\!\log _{\mathbf{u}}^r(\mathbf{v})\!=\!\frac{2}{\sqrt{r} \lambda_{\mathbf{u}}^r} \tanh^{-1}\big(\sqrt{r}\left\|-\mathbf{u} \oplus \mathbf{v}\right\|_2\big) \frac{-\mathbf{u} \oplus \mathbf{v}}{\left\|-\mathbf{u} \oplus \mathbf{v}\right\|_2}.\!\!
    \label{eq:log_map}
\end{equation}
\end{definition}

\vspace{-0.1cm}
The distance between vectors in the hyperbolic space is measured by the Riemannian distance in Definition \ref{def:Riemannian_dist}.


\vspace{-0.1cm}
\begin{definition}[\textbf{Riemannian distance}]
\label{def:Riemannian_dist}
For two points in the Poincaré ball $\mathbf{u}, \mathbf{v}\in\mathbb{D}_r^d$, their  Riemannian distance induced by the Riemannian metric $g_{\mathbb{D}}(\cdot)$ is defined as:
\vspace{-0.2cm}
\begin{equation}
	\begin{aligned}
	d_{r}(\mathbf{u}, \mathbf{v})=\frac{2}{\sqrt{r}} \tanh ^{-1}\big(\sqrt{r}{\|-\mathbf{u} \oplus \mathbf{v}\|_2}\big).
	\label{eq:riemannian_dist}
	\end{aligned}
\end{equation}
\end{definition}
\vspace{-0.2cm}

\vspace{0.1cm}
\noindent\textbf{Encoding and Projection.}  
Let $f(\mathbf{x})$ and $g(\mathbf{t}_c)$ denote ``clean'' image and text features obtained from vision and text encoders of CLIP. Let  $(\mathbf{x}\!+\!\boldsymbol{\delta}_X, \mathbf{t}_c\!+\!\boldsymbol{\delta}_T)$ be an image-text pair with perturbations $\boldsymbol{\delta}_X$ and $\boldsymbol{\delta}_T$ to  optimize.

Firstly,  clean or adversarial image/text embeddings  
are projected into the Poincaré ball $\mathbb{D}_c^d$ by the exponential map $\exp_{\mathbf{0}}^r(\cdot)$, with projection 
applied for the numerical stability: 
\vspace{-0.2cm}
\begin{equation}
	\operatorname{Proj}(\mathbf{z}):= \begin{cases}\mathbf{z} & \text { if }\|\mathbf{z}\|_2 \leq (1\!-\!\xi)\frac{1}{\sqrt{r}} \\ (1-\xi) \frac{\mathbf{z}}{\sqrt{r}\|\mathbf{z}\|_2} & \text { else }\end{cases},\!
	\label{eq:7}
\end{equation}

\noindent
where $\xi\!\geq\!0$ prevents the projected vectors from ``touching'' the Poincaré ball's boundary. Let the  image-text embeddings in the Poincaré ball be $\phi(\mathbf{x}\!+\!\boldsymbol{\delta}_X)\!=\!\operatorname{Proj}\big(\exp_{\mathbf{0}}^r(f(\mathbf{x}\!+\!\boldsymbol{\delta}_X))\big)$ and $\psi(\mathbf{t}_c\!+\!\boldsymbol{\delta}_T)\!=\!\operatorname{Proj}\big(\exp_{\mathbf{0}}^r(g(\mathbf{t}_c\!+\!\boldsymbol{\delta}_T))\big)$. 

\vspace{0.2cm}
\noindent\textbf{Hierarchical Image-text Alignment.}
For superclass text prompts, embeddings are obtained from  text CLIP. However, images do not encompass an entire superclass (universal car template does not exist). Thus,  we align a superclass with  hyperbolic average of image embeddings within a mini-batch where the base classes share the same parent.

Hyperbolic Averaging (HypAvg) in the Poincaré coordinates, based on the so-called \textit{Einstein midpoint}, is given as:
\vspace{-0.2cm}
\begin{equation}
\small
\!\!\!\!\!\mathrm{HypAvg}\big(\{\mathbf{z}_i\}_{i=1}^m\big) \!=\! \frac{\sum_{i=1}^m \left(\frac{\gamma_i}{\rho} \frac{2\mathbf{z}_i}{1 + r \|\mathbf{z}_i\|_2^2}\right)}{1 \!+\! \sqrt{1 \!-\! \frac{r}{\rho^2}\Big\lVert \sum_{i=1}^m\!\left(\gamma_i \frac{2\mathbf{z}_i}{1 + r \|\mathbf{z}_i\|_2^2}\right)\Big\rVert_2^2}}.\!\!\!
\end{equation}

\vspace{0.1cm}
\noindent
Let $\gamma_i \!=\! 1/\sqrt{1 - r\|\mathbf{k}_i\|_2^2}$  denote the Lorentz factor for each point, $\mathbf{k}_i\!=\!\frac{2\mathbf{z}_i}{1 + r \|\mathbf{z}_i\|_2^2}$ and $\rho\!=\!\sum_{i=1}^m \gamma_i$ and curvature $r$. 


For samples of base category $c$ in the set $\mathcal{X}_c$ (\eg, formed from mini-batch samples of base category $c$), one can obtain $\mathrm{HypAvg}\big(\{ \phi(\mathbf{{x}})\!:\! \mathbf{{x}} \!\in\! \mathcal{X}_c \}\big)$. Then one constructs image embeddings for the corresponding higher-level superclasses by applying hyperbolic averaging over the (average) embeddings at the lower-class level. This process is performed recursively across all hierarchical levels. Thus, for level $l\!=\!0,\ldots,L$ and category $c\!=\!1,\ldots,C_l$, we get $\boldsymbol{\phi}_c^l\!=\!\mathrm{HypAvg}\big(\{ \phi(\mathbf{{x}})\!:\! \mathbf{{x}} \!\in\! \mathcal{X}^l_c \}\big)$. By analogy, for adversarial samples, $\hat{\mathcal{X}}^l_c$, we get $\hat{\boldsymbol{\phi}}_c^l\!=\!\mathrm{HypAvg}\big(\{ \phi(\mathbf{{x}})\!:\! \mathbf{{x}} \!\in\! \hat{\mathcal{X}}^l_c \}\big)$.


\begin{table*}[!t]
	\vspace{-0.1cm}
	\centering
	\caption{Zero-shot \textbf{clean} accuracy (\%). Adversarial fine-tuning is conducted on ImageNet, followed by evaluations across 15 datasets.}
	\vspace{-0.3cm}
	\renewcommand{\arraystretch}{1}
	\resizebox{1\linewidth}{!}{
		\begin{tabular}{ccccccccccccccccc>{\columncolor{LightBlue}}c}
			\toprule
			\textbf{Method} & \negs\negsr & \rotatebox{90}{\datas{Image}{Net}} & \rotatebox{90}{STL10} & \rotatebox{90}{\datas{CIFAR}{10}} & \rotatebox{90}{\datas{CIFAR}{100}} & \rotatebox{90}{\datas{SUN}{397}} & \rotatebox{90}{\datas{Stanford}{Cars}} & \rotatebox{90}{\datas{Food}{101}} & \rotatebox{90}{\datas{Oxford}{Pet}} & \rotatebox{90}{\datas{Flower}{102}} & \rotatebox{90}{DTD} & \rotatebox{90}{\datas{Euro}{Sat}} & \rotatebox{90}{FGVC} & \rotatebox{90}{PCAM} & \rotatebox{90}{\datas{Caltech}{101}} & \rotatebox{90}{\datas{Caltech}{256}} & \textbf{Average} \\
			\midrule
			CLIP & \negs2021 \cite{radford2021learning}\negsr & 59.13 & 97.17 & 88.55 & 62.29 & 57.68 & 52.07 & 83.84 & 87.35 & 65.60 & 40.05 & 38.31 & 20.13 & 52.26 & 87.08 & 82.01 & 64.90 \\
			\midrule
			TeCoA & \negs2023 \cite{MaoGYWV23}\negsr & 58.69 & 92.15& 75.89 & 46.31 & 48.67 & 26.59 & 47.27 & 79.42 & 45.15 & 31.70 & 25.32 & 12.15 & 47.23 & 79.20 & 73.51 & 52.62 \\
			PMG-FT & \negs2024 \cite{wang2024pre}\negsr & 60.20 & 93.89 & 80.79 & 51.92 & 53.55 & 40.49 & 61.26 & 82.91 & 53.39 & 33.09 & 24.29 & 14.76 & 48.47 & 84.00 & 77.40 & 57.36 \\
			
			FARE & \negs2024 \cite{schlarmann2024robust}\negsr & 57.86 & 94.81 & 85.45 & 60.75 & 54.05 & 45.06 & 67.00 & 84.64 & 58.72 & 36.23 & 24.84 & 16.23 & 44.89 & 85.42 & 79.13 & 59.67 \\
            AoS & \negs$\!\!\!$2025 \cite{dong2025robustifying}\negsr & 60.58 & 96.83 & 86.70 & 61.98 & 55.94 & 46.42 & 69.53 & 85.80 & 59.69 & 38.06 & 29.25 & 17.00 & 50.22 & 86.60 & 80.93 & 61.70 \\
			\rowcolor{LightBlue}\textbf{Ours} & \negs\negsr & 61.19 & 96.46 & 86.50 & 62.77 & 55.76 & 48.11 & \textbf{70.31} & 85.63 & 60.32 & \textbf{38.25} & 30.41 & 17.39 & 50.90 & 86.81 & 81.27 & 62.14 \\
			\rowcolor{LightBlue}\textbf{Ours (5 trees)} & \negs\negsr & \textbf{61.86} & \textbf{97.18} & \textbf{87.13} & \textbf{62.87} & \textbf{56.29} & \textbf{48.23} & 69.76 & \textbf{86.32} & \textbf{60.82} & 37.68 & \textbf{30.93} & \textbf{17.76} & \textbf{51.69} & \textbf{86.95} & \textbf{81.85} & \textbf{62.49} \\
			\bottomrule
		\end{tabular}
	}
	\label{tab:1}
\end{table*}

\begin{table*}[!t]
	\vspace{-0.1cm}
	\centering
	\caption{Zero-shot \textbf{robust} accuracy (\%). Adversarial samples are generated by the PGD attack of $20$ steps with the image-level perturbation radius $\epsilon_{\text{X}}=1/255$. Adversarial fine-tuning is conducted on ImageNet, followed by robustness evaluations over 15 datasets.}
	\vspace{-0.3cm}
	\renewcommand{\arraystretch}{0.9}
	\resizebox{1\linewidth}{!}{
		\begin{tabular}{ccccccccccccccccc>{\columncolor{LightBlue}}c}
			\toprule
			\textbf{Method} & \negs\negsr & \rotatebox{90}{\datas{Image}{Net}} & \rotatebox{90}{STL10} & \rotatebox{90}{\datas{CIFAR}{10}} & \rotatebox{90}{\datas{CIFAR}{100}} & \rotatebox{90}{\datas{SUN}{397}} & \rotatebox{90}{\datas{Stanford}{Cars}} & \rotatebox{90}{\datas{Food}{101}} & \rotatebox{90}{\datas{Oxford}{Pet}} & \rotatebox{90}{\datas{Flower}{102}} & \rotatebox{90}{DTD} & \rotatebox{90}{\datas{Euro}{Sat}} & \rotatebox{90}{FGVC} & \rotatebox{90}{PCAM} & \rotatebox{90}{\datas{Caltech}{101}} & \rotatebox{90}{\datas{Caltech}{256}} & \textbf{Average} \\
			\midrule
			CLIP &\negs2021 \cite{radford2021learning}\negsr & 1.48 & 38.50 & 10.56 & 4.85 & 1.21 & 0.27 & 6.94 & 3.79 & 1.38 & 3.03 & 0.05 & 0.00 & 0.08 & 22.04 & 14.00 & 7.21 \\
			\midrule
			TeCoA & \negs2023 \cite{MaoGYWV23}\negsr & 41.48 & 83.50 & 60.08 & 34.16 & 31.55 & 13.08 & 27.28 & 62.80 & 28.80 & 22.71 & 16.58 & 5.88 & 26.81 & 69.18 & 59.80 & 38.91 \\
			PMG-FT & \negs2024 \cite{wang2024pre}\negsr & 38.94 & 84.00 & 62.27 & 35.92 & 31.07 & 16.74 & 31.10 & 63.07 & 31.99 & 23.14 & 14.94 & 6.06 & 26.10 & 70.85 & 59.57 & 39.72 \\
			FARE & \negs2024 \cite{schlarmann2024robust}\negsr & 29.80 & 84.40 & 65.03 & 38.98 & 25.59 & 17.44 & 32.05 & 56.70 & 29.88 & 24.05 & 10.15 & 4.30 & 22.51 & 69.40 & 58.63 & 37.93 \\
            AoS & \negs$\!\!\!$2025 \cite{dong2025robustifying}\negsr & 47.27 & 86.10 & 67.69 & 40.23 & 32.46 & 21.25 & 34.42 & 67.80 & 35.88 & 25.86 & 17.32 & 8.03 & 36.19 & 73.70 & 63.98 & 43.88 \\
			\rowcolor{LightBlue}\textbf{Ours} &\negs\negsr & 47.81 & 86.97 & 68.36 & 41.89 & 33.53 & 20.88 & 34.85 & 67.96 & 36.60 & 25.41 & 17.49 & 8.58 & 36.49 & 74.12 & 64.15 & 44.34 \\
			\rowcolor{LightBlue}\textbf{Ours (5 trees)} & \negs\negsr & \textbf{49.09} & \textbf{88.16} & \textbf{69.78} & \textbf{42.66} & \textbf{34.16} & \textbf{22.25} & \textbf{35.89} & \textbf{68.05} & \textbf{37.02} & \textbf{26.78} & \textbf{18.89} & \textbf{9.90} & \textbf{37.82} & \textbf{74.99} & \textbf{65.41} & \textbf{45.39} \\
			\bottomrule
			
		\end{tabular}
	}
	\label{tab:2}
	\vspace{-0.3cm}
\end{table*}

Thus, we propose a Hierarchy-aware Negative Set Augmentation (HNSA) integrating both the higher- and lower-level superclasses into fine-tuning. The probability of sample being class $c$ at level $l$ 
is: 
\vspace{-0.1cm}
\begin{equation}
\small
		{p}^{l}_{c} =  \frac{\operatorname{exp}\big[-\!d_{r}\big(\boldsymbol{\phi}_c^l, \psi(\mathbf{t}^l_c)\big)\,\big]}{\eta_c^l\,+\sum\limits_{c'=1}^{C_l}\operatorname{exp}\big[-\!d_{r}\big(\boldsymbol{\phi}_{c}^l, \psi(\mathbf{t}^l_{c'})\big)\,\big]},
		\label{eq:12}
\end{equation}

\vspace{-0.1cm}
\noindent
where the scalar
$\small
\eta_c^l\!\!=\!\!\sum_{k\in\mathcal{K}^{l+1}}^{C_{l+1}} \exp\Big[-\!d_{r}\big(\boldsymbol{\phi}_c^l, \psi(\mathbf{t}^{l+1}_k)\big)\Big] + \sum_{k\in\mathcal{K}^{l-1}}^{C_{l-1}} \exp\Big[-\!d_{r}\big(\boldsymbol{\phi}_c^l, \psi(\mathbf{t}^{l-1}_k)\big)\Big]$
%
is an additional augmentation with ``negative'' classes at levels $l\!-\!1$ and $l\!+\!1$. Sets $\mathcal{K}^{l-1}$ and $\mathcal{K}^{l-1}$ contain classes at levels $l\!-\!1$ and $l\!+\!1$ that do not share an edge with class $c$ at level $l$.$\!$

We define $\mathbf{{p}}^l\!=\![{p}^l_1,\ldots, {p}^l_{C_l}]^\top\!\in\![0,1]^{C_l}$ as the likelihood vector, and rewrite the standard adversary generation scheme 
from Eq. \eqref{eq:2} into its hierarchical counterpart called Hierarchy-preserving Image-Text Alignment (HITA):
\vspace{-0.2cm}
%
\begin{align}
\small
&\!\!\!\!\!\!\!\!\mathcal{L}'\!=\!\!\!\!\!\max\limits_{\substack{\left\| \boldsymbol{\delta}_{\text{X}} \right\|_{\infty}\leq\epsilon_{\text{X}}\\ \left\| \boldsymbol{\delta}_{\text{T}} \right\|_{\infty}\leq\epsilon_{\text{T}}}} \,\sum_{l=0}^{L}\omega_l\,\mathcal{L}_{\text{CE}}\Big[ \mathbf{{p}}^l\big(\big\{\mathbf{x} \!+\! \boldsymbol{\delta}_{\text{X}}\!:\!\mathbf{x}\!\in\!\mathcal{X}_c^l\big\}, \mathbf{t}_c^l \!+\! \boldsymbol{\delta}_{\text{T}}\big), \mathbf{y}(c) \Big],\nonumber\\[-12pt]
&\label{eq:13}
\end{align}

\vspace{-0.2cm}
\noindent
where $\omega_l \!=\! 1\!-\!\frac{l}{L+1}$ is a weighting factor for each hierarchical level. Universal adversarial samples for image and text modalities are generated as $\mathbf{\hat{x}}\!=\!\mathbf{x}\!+\!\boldsymbol{\delta_{\text{X}}}$ and $\mathbf{\hat{t}}\!=\!\mathbf{t}\!+\!\boldsymbol{\delta_{\text{T}}}$. To maintain the hierarchical integrity of text embeddings in the hyperbolic space, we add a soft constraint controlled by   offset $\zeta_\text{gap}\!\geq\!0$ to ensure that higher-level superclass norms are smaller than those of lower-level superclasses:
\vspace{-0.2cm}
\begin{equation}
\small
	\!\mathcal{L}^\text{Label}_\text{gap} \!=\!\! \sum_{l=0}^{L-1}\sum_{c=1}^{C_l}\max\Big(0, \big\|\psi\big(\mathbf{\hat{t}}^{l+1}_c\big)\big\|_2 - \big\|\psi\big(\mathbf{\hat{t}}^l_c\big)\big\|_2 + \zeta_\text{gap}\Big).\!\!
	\label{eq:14}
\end{equation}
\vspace{-0.3cm}

\vspace{0.1cm}
\noindent\textbf{Individual Images and Intra-class Variability.} 
\label{sec:intra_varia}
To maintain the hierarchical integrity of individual image embeddings of a given class, they cannot be fully aligned with the base class text prompt embedding as these individual images still vary at fine-grained level (no annotations are given). 
Thus, we firstly align hyperbolic image embedding $\phi(\mathbf{\hat{x}})$ to be close to the base class text embedding $\psi(\mathbf{\hat{t}}_c)$: 
%
\vspace{-0.2cm}
\begin{equation}
	\mathcal{L}_\text{vic} = \sum\nolimits_{\mathbf{\hat{x}_c}\in\mathcal{\hat{X}}_c}\Big|d_{r}\big(\phi\big(\mathbf{\hat{x}}_c\big), \psi\big(\mathbf{\hat{t}}_c\big)\big) - \zeta_\text{vic}\Big|,
	\label{eq:8}
\end{equation}

\vspace{-0.2cm}
\noindent
where $\zeta_\text{vic}\!\geq\!0$ is a radius of the text embedding. $\mathcal{\hat{X}}_c$ is the adversarial image set of the base category $c$. $\mathcal{L}_\text{vic}$ penalizes image embeddings that fall outside the radius. 

We maintain the hierarchical order of embeddings in the hyperbolic space by 
encouraging each image embedding to reside in the upper arc of its base class text embedding: 
\vspace{-0.2cm}
\begin{equation}
	\!\!\mathcal{L}^\text{Intra}_\text{gap} \!=\!\! \sum_{\mathbf{\hat{x}}_c\in\mathcal{\hat{X}}_c}\!\max\Big(0, \big\|\psi\big(\mathbf{\hat{t}}_c\big)\big\|_2 - \big\|\phi\big(\mathbf{\hat{x}_c}\big)\big\|_2 + \zeta_\text{gap}\Big),
	\label{eq:9}
\end{equation}

\vspace{-0.2cm}
\noindent
where $\zeta_\text{gap}\!\geq\!0$ controls the minimum upper arc radius. 

\vspace{0.2cm}
\noindent\textbf{Objective Function.} 
We formalize our objective   as: 
\begin{equation}
	\mathcal{L} = \mathcal{L}' + \lambda_1\,\mathcal{L}_\text{vic} + \lambda_2\, \left(\mathcal{L}^\text{Label}_\text{gap}+\mathcal{L}^\text{Intra}_\text{gap}\right),
	\label{eq:15}
\end{equation}
where $\lambda_1$ and $\lambda_2$ are tunable factors. Unlike methods  
\cite{MaoGYWV23, wang2024pre, schlarmann2024robust}, we also optimize the text encoder projection layer to enable better hierarchical alignment. 
 In  inference, we use the same evaluation protocol with previous adversarial fine-tuning approaches based on the image-text alignment.

\vspace{0.1cm}
\noindent\textbf{Theoretical Analysis.} 
The key to adversarial robustness of our approach is that we obtain more general perturbations $\boldsymbol{\delta}_X$ and $\boldsymbol{\delta}_T$ against our hierarchical classifier which enjoys several margins with size varying in $(0,\infty)$\textemdash which is not achievable in the Euclidean model (see Fig. \ref{fig:hyp_margin}).
\begin{definition}
\label{def:log-margin}
 SoftMax,   $\operatorname{exp}\big[-\!\lambda\,d_{r}(\boldsymbol{\phi}, \boldsymbol{\psi}_c)\,\big]/\big(\sum\limits_{c}\operatorname{exp}\big[-\!\lambda\,d_{r}(\boldsymbol{\phi}, \boldsymbol{\psi}_{c'})\,\big]\big)$, can be transformed  into the log margin
\vspace{-0.1cm}
\begin{align}
&\!\!\!\!\!\!-m_r^{c,c'}\!=\!\lambda\,d_{r}(\boldsymbol{\phi}, \boldsymbol{\psi}_c)+\underbrace{\log\Big(\sum\limits_{c'\neq c}\operatorname{exp}\big[-\!\lambda\,d_{r}(\boldsymbol{\phi}, \boldsymbol{\psi}_{c'})\,\big]\Big)}_{-\lambda\,\min_{c'\neq c}\,d_{r}(\boldsymbol{\phi}, \boldsymbol{\psi}_{c'}) }\nonumber\\[-14pt]
&
\end{align}

\vspace{-0.1cm}
\noindent
via $-\log(\cdot)$. Here, $\min_{c\neq c'}(\cdot)$ holds approximately, \ie, temperature  $\lambda\!>\!0$ should be sufficiently large. Then, the  log margin $m_r$ for the Riemannian distance is:
\vspace{-0.1cm}
\begin{equation}
\!\!\!\!\!\!\!\!\!\!m_{r}^{c,c'}(\eta)
\!=\!\frac{\lambda}{\sqrt{r}}
\Big[
\operatorname{acosh}\!\big(1\!+\!\alpha(\eta)\,\beta_{c'}\big)
\!-\!
\operatorname{acosh}\!\big(1\!+\!\alpha(\eta)\,\beta_c\big)
\Big],
\end{equation}
where $\Vert\boldsymbol{\phi}\rVert_2\!=\!\Vert\boldsymbol{\psi}_c\rVert_2\!=\!\eta,\forall c$ is the norm  controlling level in hierarchy for vision/text embeddings $\boldsymbol{\phi}$ and $\boldsymbol{\psi}_c$.  $\alpha(\eta)\!=\!\frac{4r\eta^2}{(1-r\eta^2)^2}$.  $\beta_c\!=\!1-\cos(\theta_c)$ \& $\beta_{c'}\!=\!1-\cos(\theta_{c'})$ are complements of cosine of angles $\theta_c\!=\!\measuredangle(\boldsymbol{\phi},\boldsymbol{\psi}_c)$ \& $\theta_{c'}\!=\!\measuredangle(\boldsymbol{\phi},\boldsymbol{\psi}_{c'})$.
\end{definition}
\begin{theorem}[Margin size trends (Fig. \ref{fig:hyp_margin})]
\label{th:margins}
Following Def. \ref{def:log-margin}, if $\eta\!\ll\!1/(2\sqrt{r})$ then $m_r$ grows linearly as $\alpha(\eta)\!\propto\!\eta^2$: the known approx. $\operatorname{acosh}(1\!+\!u)\!\approx\!\sqrt{2u}$ holds for $\alpha(\eta)\!\approx\!4r\eta^2\!\ll\!1$. Then $m_r^{c,c'}(\eta)\!
\approx\!\frac{\lambda}{\sqrt{r}} \sqrt{2\alpha(\eta)}\,\left(\sqrt{\beta_{c'}} \!-\! \sqrt{\beta_c}\right)$. For the maximum angle separation $\big(\cos(\theta_c), \cos(\theta_{c'})\big)\!=\!(1,0)$ we get $(\beta_c, \beta_{c'})\!=\!(0,1)$ yielding linear $m_r(\eta)\!\propto\!\frac{\lambda}{\sqrt{r}}\big(\sqrt{2}\big)\eta$. 

When $\eta\!\to\!1/\sqrt{r}$, $\alpha(\eta)\!\to\!\infty$: the known approximation $\operatorname{acosh}(1\!+\!u)\!\approx\! \log(2u)$ for large $u$ yields $m_r^{c,c'}\!
\approx\!\frac{\lambda}{\sqrt{r}}\log\frac{\beta_{c'}}{\beta_{c}}$ which does not depend any more on the norm $\eta$ but depends on angles $(\theta_c,\theta_{c'})$. For $(\beta_c, \beta_{c'})\!\to\!(0,1)$ one gets $m_r\!\to\!\infty$.

In contrast, under $d_{r\to0^+}(\mathbf{u},\mathbf{v})\!\propto\!\lVert\mathbf{u}\!-\!\mathbf{v}\rVert_2$ (Euclidean metric) and $
m_\text{Euc}(\eta)
= 2\lambda \eta^2 \big(\beta_{c'}\!-\!\beta_{c}\big)$ so for $(\beta_c, \beta_{c'})\!\to\!(0,1)$ one gets $m_\text{Euc}(\eta)\!=\!2\lambda \eta^2$.
\end{theorem}

\vspace{-0.5cm}
\begin{tcolorbox}[width=1.0\linewidth, colframe=blackish, colback=beaublue, boxsep=0mm, arc=2mm, left=2mm, right=2mm, top=2mm, bottom=2mm]
\textbf{Discussion.} Theorem \ref{th:margins} and Fig. \ref{fig:hyp_margin} show that
the maximum viable margin size
in hyperbolic space ranges in $(0, \infty)$ \wrt feature norm $\eta\!\to\!1/\sqrt{r}\!=\!1$ (unlike Euclidean classifier's range (0,1)). We deploy image/text features in hyperbolic space by ordering them hierarchically from generic to specialized\textemdash norms of parents are lower than norms of children. Thus, several levels of hierarchy enjoy different limits on margin sizes, which are known to control classifier generalization. As adversarial perturbations $\boldsymbol{\delta}_X$ and $\boldsymbol{\delta}_T$ encounter several  margins of different sizes during the projected gradient ascent, they become more universal. It is easier to increase parent classifier's adversarial risk due to its smaller margin size, which yields more universal perturbations due to more universal parent's label space. On children's classifier the same perturbation becomes specialized, gaining general-to-specialized robustification capability.$\!\!$
\vspace{-0.1cm}
\end{tcolorbox}

\section{Experiments}

\vspace{0.1cm}
\noindent\textbf{Datasets.}
Following previous works \cite{MaoGYWV23, wang2024pre}, we adversarially fine-tune CLIP using the ImageNet training set \cite{deng2009imagenet}. We then evaluate the fine-tuned CLIP model on the ImageNet test set and an additional $14$ zero-shot datasets spanning various image recognition tasks (details in Appendix \ref{supp:dataset_details}).

\vspace{0.1cm}
\noindent\textbf{Implementation.} 
Unless specified otherwise, we employ  CLIP  \cite{radford2021learning} with the ViT-Base/32 architecture \cite{DosovitskiyB0WZ21}, consistent with previous works \cite{MaoGYWV23, wang2024pre}. Superclasses are constructed using the ImageNet hierarchy up to depth $L=5$ for both image and text modalities. When using hierarchical forests (\eg, 2-5 category trees), the loss function in Eq. \eqref{eq:15} is applied per tree, and resulting losses are added.

During training, we generate adversarial samples using PGD \cite{MadryMSTV18} for images and texts of diverse abstraction levels, performing $3$ iterations under the $\ell_\infty$-norm threat model. For adversarial images, we set the perturbation radius and step size to $\epsilon_{\text{X}}=\alpha_{\text{X}}=1/255$, unless noted otherwise. For text-level perturbations, we set the maximum perturbation radius to $\epsilon_{\text{X}}=2\times10^{-4}$ and the step size to $\alpha_{\text{T}}=1\times10^{-4}$. During testing, in addition to natural performance, we evaluate our model against three strong white-box adversarial attacks: PGD \cite{MadryMSTV18}, CW \cite{carlini2017towards}, and Auto-Attack \cite{croce2020reliable}. 
All evaluations are conducted under adaptive attack schemes for fairness. More implementation details are in Appendix \ref{supp:implement_details}.

\begin{table}[!t]
	\vspace{-0.1cm}
	\centering
	\renewcommand{\arraystretch}{0.8}
	\caption{Average PGD-20 robust accuracy (\%) over 15 datasets across a range of perturbation radii during \textbf{evaluation only}.}
	\vspace{-0.3cm}
	\resizebox{0.8\linewidth}{!}{
		\begin{tabular}{clcccc}
			\toprule
			\multirow{2}{*}{Method} &\negsr\negsr& \multicolumn{4}{c}{PGD-20 Robust Accuracy} \\
			\cmidrule(l){3-6}
			&\negsr\negsr & 1/255 & 2/255 & 3/255 & 4/255 \\
			\midrule
			TeCoA & \negsr\cite{MaoGYWV23}\negsr & 38.91 & 25.43 & 14.72 & 8.38 \\
			PMG-FT &\negsr\cite{wang2024pre}\negsr & 39.72 & 23.38 & 12.70 & 6.58 \\
			FARE & \negsr\cite{schlarmann2024robust}\negsr & 37.93 & 24.87 & 13.37 & 7.74 \\
            AoS & \negsr\cite{dong2025robustifying}\negsr & 43.88 & 26.70 & 15.83 & 9.51 \\
			\rowcolor{LightBlue}\textbf{Ours}& \negsr\negsr & \textbf{44.34} & \textbf{27.19} & \textbf{16.05} & \textbf{9.92} \\
			
			\bottomrule
		\end{tabular}
	}
	\label{tab:3}
\end{table}

\begin{table}[!t]
	\vspace{-0.1cm}
	\centering
	\renewcommand{\arraystretch}{0.8}
	\caption{Average performance (\%) using adversaries of varying $\epsilon_{\text{X}}$ for both \textbf{adversarial fine-tuning and robustness evaluations}.}
	\vspace{-0.3cm}
	\resizebox{0.95\linewidth}{!}{
		\begin{tabular}{cclcccc}
			\toprule
			Radius $\epsilon$ & Method &\negsr\negsr & Clean & PGD & CW & AA \\
			\midrule
			\multirow{5}{*}{2/255} & TeCoA &\negsr\cite{MaoGYWV23}\negsr & 49.10 & 26.86 & 26.07 & 25.33 \\
			& PMG-FT &\negsr\cite{wang2024pre}\negsr & 50.72 & 29.38 & 28.41 & 27.66 \\
			& FARE &\negsr\cite{schlarmann2024robust}\negsr & 51.09 & 28.55 & 27.79 & 27.18 \\
            & AoS &\negsr\cite{dong2025robustifying}\negsr & 51.86 & 30.48 & 29.45 & 28.70 \\
			\rowcolor{LightBlue}\cellcolor{white}& \textbf{Ours} &\negs\negsr & \textbf{52.27} & \textbf{31.64} & \textbf{30.55} & \textbf{29.49} \\
			\midrule
			\multirow{5}{*}{3/255} & TeCoA &\negsr\cite{MaoGYWV23}\negsr & 42.90 & 19.59 & 18.69 & 17.94 \\
			& PMG-FT &\negsr\cite{wang2024pre}\negsr & 43.09 & 22.56 & 20.61 & 18.12 \\
			& FARE &\negsr\cite{schlarmann2024robust}\negsr & 43.24 & 22.30 & 20.85 & 18.32 \\
            & AoS &\negsr\cite{dong2025robustifying}\negsr & 43.98 & 23.64 & 21.78 & 19.16 \\
			\rowcolor{LightBlue}\cellcolor{white}& \textbf{Ours} & \negsr\negsr& \textbf{44.30} & \textbf{24.57} & \textbf{22.60} & \textbf{19.98} \\
			\midrule
			\multirow{5}{*}{4/255} & TeCoA &\negsr\cite{MaoGYWV23}\negsr & 37.67 & 14.80 & 13.75 & 12.96 \\
			& PMG-FT &\negsr\cite{wang2024pre}\negsr & 37.84 & 17.03 & 15.18 & 13.29 \\
			& FARE &\negsr\cite{schlarmann2024robust}\negsr & 37.95 & 16.57 & 14.21 & 13.43 \\
            & AoS &\negsr\cite{dong2025robustifying} \negsr & 38.49 & 18.38 & 16.40 & 14.05 \\
			\rowcolor{LightBlue}\cellcolor{white}& \textbf{Ours} &\negsr\negsr& \textbf{38.75} & \textbf{19.04} & \textbf{16.92} & \textbf{14.71} \\
			\bottomrule
		\end{tabular}
	}
	\label{tab:4}
	\vspace{-0.3cm}
\end{table}

\noindent\textbf{Benchmarking on 15 datasets.} 
Tables \ref{tab:1} \& \ref{tab:2} present a comparison of our method with state-of-the-art adversarial fine-tuning techniques. Apart from ImageNet, we perform zero-shot inference on 14 additional datasets, reporting accuracy on both clean images and their adversaries generated via 20-step PGD attacks. Our method consistently achieves higher clean accuracy across all datasets, with an average improvement of 2.5\% over FARE \cite{schlarmann2024robust}, narrowing the gap with standard CLIP (Table \ref{tab:1}) as expected. While the standard CLIP has limited robustness (Table \ref{tab:2}), our method enjoys a mean gain of 6.4\% over FARE in robustness. With 5 class tress, we also surpass recent AoS \cite{dong2025robustifying} by 1\% and 1.5\% on clean/robust accuracies despite AoS uses 10x more image and 50x text augmentations for subspaces.

\vspace{0.1cm}
\noindent\textbf{Robustness across multiple attack intensities.}
Beyond evaluations with adversaries of a small attack strength ($\epsilon_{\text{X}}=1/255$), we assess the zero-shot adversarial robustness under larger perturbation radii (stronger attack intensities). Table \ref{tab:3} shows that our method consistently outperforms other approaches across a spectrum of attack intensities.

\vspace{0.1cm}
\noindent\textbf{Fine-tuning with diverse perturbation radii.}
Unlike previous studies \cite{MaoGYWV23, wang2024pre} that focus on a small perturbation radius of $\epsilon_{\text{X}}=1/255$ during fine-tuning, we explore larger $\ell_{\infty}$-norm perturbation radii of $\epsilon_{\text{X}} = 2/255$, $3/255$, and $4/255$. For fair comparisons, fine-tuned models are evaluated against adversaries generated with \textit{matching perturbation levels}. Table \ref{tab:4} shows that our method outperforms  prior approaches on both clean and adversarial samples  
under fine-tuning with larger perturbation radii.

\begin{table}[!t]
	\vspace{-0.1cm}
	\centering
	\renewcommand{\arraystretch}{0.8}
	\caption{Average performance (\%) across different vision architectures of CLIP using the perturbation radius of $\epsilon_{\text{X}}=1/255$.}
	\vspace{-0.3cm}
	\resizebox{1\linewidth}{!}{
		\begin{tabular}{cclcccc}
			\toprule
			Architecture & Method &\negsr\negsr & Clean & PGD & CW & AA \\
			\midrule
			\multirow{5}{*}{ViT-B} & TeCoA &\negsr\cite{MaoGYWV23}\negsr & 52.62 & 38.91 & 37.85 & 37.62 \\
			&PMG-FT &\negsr\cite{wang2024pre}\negsr & 57.36 & 39.72 & 38.70 & 38.09 \\
			&FARE &\negsr\cite{schlarmann2024robust}\negsr & 59.67 & 37.93 & 37.56 & 37.18 \\
            &AoS &\negsr\cite{dong2025robustifying}\negsr & 61.70 & 43.88 & 42.94 & 42.18 \\
			&\cellcolor{LightBlue}\textbf{Ours} &\cellcolor{LightBlue}\negsr\negsr& \cellcolor{LightBlue}\textbf{62.14} & \cellcolor{LightBlue}\textbf{44.34} & \cellcolor{LightBlue}\textbf{43.58} & \cellcolor{LightBlue}\textbf{42.72} \\
			
			\midrule
			
			\multirow{5}{*}{ViT-L} & TeCoA &\negsr\cite{MaoGYWV23}\negsr & 66.39 & 42.86 & 39.08 & 38.43 \\
			&PMG-FT &\negsr\cite{wang2024pre}\negsr & 67.11 & 43.64 & 39.56 & 38.91 \\
			&FARE &\negsr\cite{schlarmann2024robust}\negsr & 67.71 & 43.18 & 40.23 & 39.62 \\
            &AoS &\negsr\cite{dong2025robustifying}\negsr & \textbf{68.41} & 45.76 & 44.13 & 43.49 \\
			&\cellcolor{LightBlue}\textbf{Ours} &\cellcolor{LightBlue}\negsr\negsr& \cellcolor{LightBlue}68.38 & \cellcolor{LightBlue}\textbf{46.40} & \cellcolor{LightBlue}\textbf{44.97} & \cellcolor{LightBlue}\textbf{44.25} \\
			
			\midrule
			
			\multirow{5}{*}{ResNet-50} & TeCoA &\negsr\cite{MaoGYWV23}\negsr & 42.12 & 28.45 & 27.72 & 27.13 \\
			&PMG-FT &\negsr\cite{wang2024pre}\negsr & 46.03 & 30.66 & 29.20 & 28.36 \\
			&FARE &\negsr\cite{schlarmann2024robust}\negsr & 48.53 & 29.16 & 28.41 & 27.83 \\
            &AoS &\negsr\cite{dong2025robustifying}\negsr & 49.60 & 32.95 & 32.14 & 31.57 \\
			&\cellcolor{LightBlue}\textbf{Ours} &\cellcolor{LightBlue}\negsr\negsr& \cellcolor{LightBlue}\textbf{49.87} & \cellcolor{LightBlue}\textbf{33.29} & \cellcolor{LightBlue}\textbf{32.50} & \cellcolor{LightBlue}\textbf{31.98} \\
			
			\bottomrule
		\end{tabular}
	}
	\label{tab:5}
    \vspace{-0.3cm}
\end{table}

\vspace{0.1cm}
\noindent\textbf{Robustness on different architectures.}
Below we experiment with architectures other than CLIP with a ViT-B backbone, specifically ViT-L and ResNet-50. Table \ref{tab:5} shows that our approach consistently surpasses previous adversarial fine-tuning methods 
across 15 datasets with larger backbones yielding better clean and adv. robustness results.

\begin{table}[!t]
	\vspace{-0.2cm}
	\centering
	\renewcommand{\arraystretch}{0.8}
	\caption{Average performance (\%) using adversaries of varying $\epsilon_{\text{X}}$ for both fine-tuning and robustness evaluations with \textbf{VPT}.}
	\vspace{-0.3cm}
	\resizebox{0.95\linewidth}{!}{
		\begin{tabular}{cclcccc}
			\toprule
			Radius $\epsilon$ & Method &\negsr\negsr& Clean & PGD & CW & AA \\
			\midrule
			\multirow{5}{*}{1/255} & TeCoA &\negsr\cite{MaoGYWV23}\negsr & 51.00 & 32.27 & 31.11 & 30.26 \\
			& PMG-FT &\negsr\cite{wang2024pre}\negsr & 52.64 & 33.09 & 32.10 & 30.83 \\
			& FARE &\negsr\cite{schlarmann2024robust}\negsr & 52.75 & 32.69 & 31.58 & 30.64 \\
            & AoS &\negsr\cite{dong2025robustifying}\negsr & 54.43 & 34.38 & 33.27 & 32.05 \\
			& \cellcolor{LightBlue}\textbf{Ours} &\cellcolor{LightBlue}\negsr\negsr& \cellcolor{LightBlue}\textbf{54.70} & \cellcolor{LightBlue}\textbf{34.97} & \cellcolor{LightBlue}\textbf{33.56} & \cellcolor{LightBlue}\textbf{32.29} \\
			\midrule
			\multirow{5}{*}{2/255} & TeCoA &\negsr\cite{MaoGYWV23}\negsr & 42.61 & 18.12 & 16.88 & 15.39 \\
			& PMG-FT &\negsr\cite{wang2024pre}\negsr & 42.11 & 19.26 & 17.68 & 16.47 \\
			& FARE &\negsr\cite{schlarmann2024robust}\negsr & 42.81 & 18.98 & 17.46 & 16.35 \\
            & AoS &\negsr\cite{dong2025robustifying}\negsr & 43.84 & 20.47 & \textbf{18.60} & 17.29 \\
			& \cellcolor{LightBlue}\textbf{Ours} &\cellcolor{LightBlue}\negsr\negsr & \cellcolor{LightBlue}\textbf{44.13} & \cellcolor{LightBlue}\textbf{20.65} & \cellcolor{LightBlue}18.49 & \cellcolor{LightBlue}\textbf{17.54} \\
			\midrule
			\multirow{5}{*}{3/255} & TeCoA &\negsr\cite{MaoGYWV23}\negsr & 33.86 & 12.32 & 10.78 & 8.89 \\
			& PMG-FT &\negsr\cite{wang2024pre}\negsr & 32.52 & 12.87 & 11.36 & 9.38 \\
			& FARE &\negsr\cite{schlarmann2024robust}\negsr & 33.70 & 12.47 & 10.92 & 9.04 \\
            & AoS &\negsr\cite{dong2025robustifying}\negsr & 35.13 & 13.79 & 12.42 & 10.14 \\
			& \cellcolor{LightBlue}\textbf{Ours} &\cellcolor{LightBlue}\negsr\negsr& \cellcolor{LightBlue}\textbf{35.38} & \cellcolor{LightBlue}\textbf{14.30} & \cellcolor{LightBlue}\textbf{12.81} & \cellcolor{LightBlue}\textbf{10.49} \\
			\midrule
			\multirow{5}{*}{4/255} & TeCoA &\negsr\cite{MaoGYWV23}\negsr & 26.78 & 11.04 & 9.87 & 7.19 \\
			& PMG-FT &\negsr\cite{wang2024pre}\negsr & 23.57 & 11.73 & 10.01 & 7.26 \\
			& FARE &\negsr\cite{schlarmann2024robust}\negsr & 26.17 & 11.49 & 10.32 & 7.53 \\
            & AoS &\negsr\cite{dong2025robustifying}\negsr & 27.92 & 13.17 & 11.50 & 8.87 \\
			&\cellcolor{LightBlue}\textbf{Ours} &\cellcolor{LightBlue}\negsr\negsr& \cellcolor{LightBlue}\textbf{28.40} & \cellcolor{LightBlue}\textbf{13.54} & \cellcolor{LightBlue}\textbf{11.97} & \cellcolor{LightBlue}\textbf{9.28} \\
			\bottomrule
		\end{tabular}
	}
	\label{tab:6}
\end{table}

\vspace{0.1cm}
\noindent\textbf{Adversarial fine-tuning with parameter-efficient VPT.}
Adversarial fine-tuning on all parameters of the image encoder is computationally costly. 
Thus, we employ Visual Prompt Tuning (VPT) \cite{jia2022visual}, a parameter-efficient strategy that introduces a minimal number of learnable parameters at the token embedding level. 
Table \ref{tab:6} shows that we achieve superior zero-shot accuracy (clean/robust) using robust VPT models with various configurations, trained and tested using the same perturbation radius for consistency. 

\begin{table}[!t]
	\vspace{-0.1cm}
	\centering
	\renewcommand{\arraystretch}{1}
	\caption{Average robust accuracy (\%) of diverse adversarial fine-tuning methods against text-level and bi-level adversarial attacks.}
	\vspace{-0.3cm}
	\resizebox{0.9\linewidth}{!}{
		\begin{tabular}{clcccc}
			\toprule
			\multirow{2}{*}{Method} &\multirow{2}{*}{\negsr\negsr}& \multicolumn{2}{c}{Text-Level Attacks} & \multicolumn{2}{c}{Bi-Level Attacks} \\
			\cmidrule(l){3-6}
			&\negsr\negs& \negsr\negsr BERT-Attack & GBDA & Co-Attack & SGA \\
			\midrule
			TeCoA &\negsr\cite{MaoGYWV23}\negsr & 36.22 & 34.97 & 25.87 & 25.14 \\
			PMG-FT &\negsr\cite{wang2024pre}\negsr & 37.25 & 36.73 & 26.95 & 26.76 \\
			FARE &\negsr\cite{schlarmann2024robust}\negsr & 35.76 & 35.08 & 25.10 & 24.87 \\
            AoS &\negsr\cite{dong2025robustifying}\negsr & 40.83 & 40.37 & 30.24 & 29.51 \\
			\rowcolor{LightBlue}\textbf{Ours} &\negsr\negsr& \textbf{41.78} & \textbf{41.55} & \textbf{31.76} & \textbf{31.13} \\
			
			\bottomrule
		\end{tabular}
	}
	\label{tab:7}
	\vspace{-0.3cm}
\end{table}

\vspace{0.1cm}
\noindent\textbf{Robustness to text-level and bi-level adv. attacks.}
Below we include text- and bi-level adversarial attacks. 
We report: (i) \textit{text-level adversarial attacks} by BERT-Attack \cite{LiMGXQ20}, (ii) Gradient-Based Distributional Attack (GBDA) \cite{GuoSJK21},  (iii) \textit{bi-level adversarial attacks} using Collaborative Multimodal Adversarial Attack (Co-Attack) \cite{zhang2022towards} and (iv) Set-level Guidance Attack (SGA) \cite{lu2023set}. Table \ref{tab:7} (robust accuracy) shows that our method significantly boosts robustness against competitors on  text-level and bi-level attacks.


\vspace{0.1cm}
\noindent\textbf{Extension to BLIP/CoCa.} 
We extend our model to BLIP \cite{li2022blip}, and CoCa \cite{YuWVYSW22} for vision-language understanding tasks such as image-text retrieval and image captioning. Following the settings from Appendix \ref{supp:implement_details}, we report zero-shot performance (clean and adversarial) from the Flickr30k dataset \cite{plummer2015flickr30k} for image-text retrieval and the Nocaps dataset \cite{agrawal2019nocaps} for image captioning. Table \ref{tab:8} shows that our method has consistently best results across diverse tasks.

\begin{table}[!t]
	\vspace{-0.2cm}
	\centering
	\renewcommand{\arraystretch}{1}
	\caption{\jh{Extension to BLIP/CoCa evaluated on clean samples and PGD-20 adversaries in image-text retrieval and image captioning.}}
	\vspace{-0.3cm}
	\resizebox{1\linewidth}{!}{
		\begin{tabular}{cclcccccc}
			\toprule 
			\multirow{2}{*}{\makecell{\\Architecture}} & \multirow{2}{*}{\makecell{\\Method}}\negsr\negsr\negsr &\negsr\negsr& \multicolumn{4}{c}{\textbf{Image-Text Retrieval}} & \multicolumn{2}{c}{\textbf{Image Captioning}} \\
			\cmidrule(lr){4-7} \cmidrule(lr){8-9}
			&&& \makecell{Clean\\TR} & \makecell{Robust\\TR} & \makecell{Clean\\IR} & \makecell{Robust\\IR} & \makecell{Clean\\CIDEr} & \makecell{Robust\\CIDEr} \\
			\midrule
			\multirow{5}{*}{\textbf{BLIP}} & TeCoA &\negs$\quad$\cite{MaoGYWV23}\negsr & 87.5 & 54.4 & 77.0 & 47.5 & 96.9 & 57.8 \\
			& PMG-FT &\negs$\quad$\cite{wang2024pre}\negsr & 87.8 & 55.6 & 77.9 & 48.2 & 97.5 & 58.2 \\
			&FARE &\negs$\quad$\cite{schlarmann2024robust}\negsr & 88.2 & 55.9 & 78.4 & 49.0 & 98.1 & 58.7 \\
			&\cellcolor{LightBlue}\textbf{Ours}\negsr\negsr &\cellcolor{LightBlue}\negsr\negsr& \cellcolor{LightBlue}{90.8} & \cellcolor{LightBlue}57.9 & \cellcolor{LightBlue}{80.2} & \cellcolor{LightBlue}51.4 & \cellcolor{LightBlue}101.0 & \cellcolor{LightBlue}62.5 \\
			&\multicolumn{2}{c}{\cellcolor{LightBlue}$\!\!\!\!\!\!\!\!\!\!$\textbf{Ours (2 trees)}\negsr\negsr\negsr} & \cellcolor{LightBlue}\textbf{90.9} & \cellcolor{LightBlue}\textbf{58.3} & \cellcolor{LightBlue}\textbf{80.4} & \cellcolor{LightBlue}\textbf{51.9} & \cellcolor{LightBlue}\textbf{101.4} & \cellcolor{LightBlue}\textbf{63.0} \\
			\midrule
			\multirow{5}{*}{\textbf{CoCa}} & TeCoA &\negs$\quad$\cite{MaoGYWV23}\negsr & 88.4 & 56.5 & 76.8 & 48.3 & 100.5 & 59.2 \\
			& PMG-FT &\negs$\quad$\cite{wang2024pre}\negsr & 88.2 & 57.3 & 76.3 & 49.7 & 102.1 & 60.2 \\
			&FARE &\negs$\quad$\cite{schlarmann2024robust}\negsr & 89.5 & 57.0 & 77.5 & 49.8 & 101.8 & 60.5 \\
			&\cellcolor{LightBlue}\textbf{Ours}\negsr\negsr &\cellcolor{LightBlue}& \cellcolor{LightBlue}91.9 & \cellcolor{LightBlue}59.4 & \cellcolor{LightBlue}79.6 & \cellcolor{LightBlue}52.6 & \cellcolor{LightBlue}105.4 & \cellcolor{LightBlue}63.8 \\
			&\multicolumn{2}{c}{\cellcolor{LightBlue}$\!\!\!\!\!\!\!\!\!\!$\textbf{Ours (2 trees)}\negsr\negsr\negsr} &\cellcolor{LightBlue}\textbf{92.3} & \cellcolor{LightBlue}\textbf{60.2} & \cellcolor{LightBlue}\textbf{79.8} & \cellcolor{LightBlue}\textbf{53.5} & \cellcolor{LightBlue}\textbf{105.7} & \cellcolor{LightBlue}\textbf{64.6} \\
			\bottomrule
		\end{tabular}
	}
	\label{tab:8}
\end{table}

\begin{table}[!t]
	\vspace{-0.1cm}
	\centering
	\renewcommand{\arraystretch}{1}
	\caption{\jh{Extensions to the medical CLIP on clean and PGD-20 adversarial samples evaluated by the AUC score.}}
	\vspace{-0.3cm}
	\resizebox{1\linewidth}{!}{
		\begin{tabular}{clcccccc}
			\toprule
			\multirow{2}{*}{Method}\negsr\negsr&\negsr\negsr & \multicolumn{2}{c}{ChestXray14} & \multicolumn{2}{c}{CheXpert} & \multicolumn{2}{c}{PadChest} \\
			\cmidrule(l){3-4} \cmidrule(l){5-6} \cmidrule(l){7-8}
			&\negsr\negsr& Clean & PGD & Clean & PGD & Clean & PGD \\
			\midrule
			TeCoA &\negsr\cite{MaoGYWV23}\negsr & 0.674 & 0.526 & 0.857 & 0.685 & 0.602 & 0.483 \\
			PMG-FT &\negsr\cite{wang2024pre}\negsr & 0.692 & 0.538 & 0.850 & 0.688 & 0.619 & 0.495 \\
            FARE &\negsr\cite{schlarmann2024robust}\negsr & 0.687 & 0.533 & 0.845 & 0.679 & 0.615 & 0.490 \\
            AoS &\negsr\cite{dong2025robustifying}\negsr & 0.718 & 0.572 & 0.883 & 0.719 & 0.643 & 0.531 \\
			\rowcolor{LightBlue}\textbf{Ours}&\negsr\negsr & 0.727 & 0.563 & 0.874 & 0.716 & 0.637 & 0.554 \\
\rowcolor{LightBlue}\multicolumn{2}{c}{\cellcolor{LightBlue}$\!\!\!\!\!\!\!\!\!\!$\textbf{Ours (2 trees)}\negsr\negsr\negsr} &\textbf{0.730} & \textbf{0.571} & \textbf{0.876} & \textbf{0.724} & \textbf{0.639} & \textbf{0.562} \\
			
			\bottomrule
		\end{tabular}
 	}
	\label{tab:9}
	\vspace{-0.3cm}
\end{table}

\vspace{0.1cm}
\noindent\textbf{Extension to Medical CLIP.} 
Below we apply our method to medical imaging, addressing adversarial vulnerabilities in computer-aided diagnosis \cite{zhao2023clip}. Following the experimental protocols in Appendix \ref{supp:implement_details}, we assess the zero-shot adversarial robustness of chest X-rays with multi-label disease annotations. Table \ref{tab:9} shows our approach consistently achieves superior AUC  for clean samples and adversaries across various radiology datasets. Even on the challenging PadChest dataset, which includes a long-tail distribution of 192 diseases, 
our method exhibits notable robustness. 

\definecolor{ao(english)}{rgb}{0.0, 0.0, 0.5}

\subsection{Analysis}

\noindent\textbf{Impact of loss components.} 
$\;$We study: 
(i) Hierarchy-pre-

\setlength{\columnsep}{8pt}%
\begin{wraptable}{r}{0.24\textwidth}
\vspace{-0.3cm}
\caption{Ablation study of three loss components in our method for average clean and robust accuracy (\%) on 15 datasets.}
\label{tab:10}
\vspace{-0.3cm}
\resizebox{0.24\textwidth}{!}
{\renewcommand{\arraystretch}{1}
		\begin{tabular}{c@{\hspace{-0.05em}}c@{\hspace{0.5em}}c@{\hspace{0.5em}}c@{\hspace{0.5em}}ccc}
			\toprule
			& HITA & vic & des & Clean & PGD & AA \\
			\midrule
			1 & & & & 52.62 & 38.91 & 37.62 \\
			2 & \ding{51} & & & 59.27 & 42.08 & 40.59 \\
			3 & \ding{51} & \ding{51} & & \textbf{62.35} & 43.67 & 42.18 \\
			4 & \ding{51} & & \ding{51} & 60.06 & 43.15 & 41.54 \\
			\midrule
			\rowcolor{LightBlue} 5 & \ding{51} & \ding{51} & \ding{51} & \textbf{62.14} & \textbf{44.34} & \textbf{42.72} \\
			\bottomrule
		\end{tabular}
}
\vspace{-0.3cm}
\end{wraptable}

\noindent
serving  Image-Text Alignment (HITA) $\mathcal{L}'$ from Eq. (\ref{eq:13}), (ii) intra-class vicinity alignment $\mathcal{L}_\text{vic}$ in Eq. (\ref{eq:8}), and (iii) intra-class and label  norm gap penalties $\mathcal{L}_\text{gap}^\text{Intra}$ \& $\mathcal{L}_\text{gap}^\text{Label}$ from Eq. (\ref{eq:9}) \& (\ref{eq:14}). 
 Table \ref{tab:10} reports the average zero-shot accuracy (clean and adversarial) on 15 datasets. Our baseline (top row) is based on TeCoA \cite{MaoGYWV23}. Incorporating hierarchy-level image-text alignment  boosts  clean/robust accuracy due to rich hierarchical decision boundary. The vicinity/intra-class penalties mitigate image embedding collapse on text embeddings. The label norm gap  promotes the hierarchical integrity.

\begin{table}[!t]
	\vspace{-0.1cm}
	\centering
	\renewcommand{\arraystretch}{1}
	\caption{Performance (\%) of diverse negative set augmentations.}
	\vspace{-0.3cm}
	\resizebox{0.9\linewidth}{!}{
		\begin{tabular}{cccc}
			\toprule
			\makecell{Negative Set Augmentation Strategy} & Clean & PGD & AA  \\
			\midrule
			w/o augmentation & 60.88 & 43.16 & 41.47 \\
			Lower-level superclasses & 61.25 & 43.62 & 41.93 \\
			Higher-level superclasses & 61.70 & 43.83 & 42.05 \\
			\rowcolor{LightBlue}Higher- \& lower-level superclasses & \textbf{62.14} & \textbf{44.34} & \textbf{42.72} \\
			\bottomrule
		\end{tabular}
	}
	\label{tab:12}
\end{table}

\vspace{0.1cm}
\noindent\textbf{Different negative set augmentation strategies.} 
Recall that we design a hierarchy-aware negative augmentation scheme in Eq. (\ref{eq:12}) based on the hyperbolic embeddings from both higher- and lower-level superclasses. 
Table \ref{tab:12} ablates different negative set augmentation strategies, and shows that integrating superclasses at hierarchical levels up+below the positive class works the best.

\begin{table}[!t]
	\vspace{-0.1cm}
	\centering
	\renewcommand{\arraystretch}{1}
	\caption{Performance (\%) of diverse adversary generation strategies for adv. fine-tuning with the average training time per epoch.}
	\vspace{-0.3cm}
	\resizebox{\linewidth}{!}{
		\begin{tabular}{ccccc}
			\toprule
			\makecell{Adversary Generation Strategy} & Clean & PGD & AA & Time (min) \\
			\midrule
			Leaf-level disruption only & 60.25 & 41.59 & 40.23 & 70.1 \\
			Diverse hierarchical perturbations & 61.37 & 42.35 & 40.88 & 154.2 \\
			\rowcolor{LightBlue}\textbf{Universal hierarchical perturbation} & \textbf{62.14} & \textbf{44.34} & \textbf{42.72} & 73.2 \\
			\bottomrule
		\end{tabular}
	}
	\label{tab:13}
\end{table}

\vspace{0.1cm}
\noindent\textbf{Different adversary generation strategies.}
Below we explore three adversary generation strategies:
\begin{itemize}
	\item \textbf{Leaf-level disruption only}: Adversarial perturbations target solely the base category classification.
	\item \textbf{Diverse hierarchical perturbations}: Perturbations are crafted at each hierarchy level for all original samples.
	\item \textbf{Universal hierarchical perturbation}: A single perturbation is applied universally across all hierarchy levels.
\end{itemize}
Table \ref{tab:13} reports zero-shot clean and robust accuracy with the training time. The universal strategy offers a favorable trade-off between robustness and computational efficiency. During testing, all methods enjoy the same inference time. 

\begin{table}[!t]
	\vspace{-0.1cm}
	\centering
	\renewcommand{\arraystretch}{1}
	\caption{\jh{Performance (\%) of LLMs for superclass generation.}}
	\vspace{-0.3cm}
	\resizebox{0.95\linewidth}{!}{
		\begin{tabular}{cccccccc}
			\toprule
			LLM for & \multirow{2}{*}{Clean} & \multicolumn{2}{c}{Image Attack} & \multicolumn{2}{c}{Text Attack} & \multicolumn{2}{c}{VLM Attack}\\
			\cmidrule(r){3-4} \cmidrule(r){5-6} \cmidrule(r){7-8}
			Superclass Generation&&PGD&AA&BERT-Attack&GBDA&CoAttack&SGA\\
			\midrule
			LLaMA-2 & 61.87 & 43.55 & 42.05 & 40.91 & 40.72 & 31.15 & 30.48 \\
			Claude-2 & 61.83 & 43.79 & 42.32 & 41.35 & 41.11 & 31.60 & 30.86 \\
			\rowcolor{LightBlue}\textbf{ChatGPT-4o} & \textbf{62.14} & \textbf{44.34} & \textbf{42.72} & \textbf{41.78} & \textbf{41.55} & \textbf{31.76} & \textbf{31.13} \\
			\bottomrule
		\end{tabular}
	}
	\label{tab:LLM_Sup_Gen}
\end{table}

\vspace{0.1cm}
\noindent\textbf{\jh{Different LLMs in superclass generation.}}
Recall that we generate superclasses using  ChatGPT-4o \cite{openai_chatgpt_2024} in the absence of predefined hierarchical taxonomies in datasets other than ImageNet.  Table \ref{tab:LLM_Sup_Gen} investigates superclass generation with powerful LLMs (LLaMA-2 \cite{touvron2023llama}, Claude-2 \cite{anthropic2023claude2}) and shows that our method is robust to these diverse choices. 

\begin{table}[!t]
	\vspace{-0.1cm}
	\centering
	\renewcommand{\arraystretch}{1}
	\caption{\jh{Performance (\%) of different  BLIP adaptation strategies.}}
	\vspace{-0.3cm}
	\resizebox{1\linewidth}{!}{
		\begin{tabular}{cccccccc}
			\toprule 
			\multirow{2}{*}{\makecell{\\Strategy}} & \multirow{2}{*}{\makecell{\\Method}} & \multicolumn{4}{c}{\textbf{Image-Text Retrieval}} & \multicolumn{2}{c}{\textbf{Image Captioning}} \\
			\cmidrule(lr){3-6} \cmidrule(lr){7-8}
			&& \makecell{Clean\\TR} & \makecell{Robust\\TR} & \makecell{Clean\\IR} & \makecell{Robust\\IR} & \makecell{Clean\\CIDEr} & \makecell{Robust\\CIDEr} \\
			\midrule
			\multirow{2}{*}{{Fine-Tuning Target VLMs}} &FARE \cite{schlarmann2024robust} & 88.2 & 55.9 & 78.4 & 49.0 & 98.1 & 58.7 \\
			&\cellcolor{LightBlue}\textbf{Ours} & \cellcolor{LightBlue}\textbf{90.8} & \cellcolor{LightBlue}\textbf{57.9} & \cellcolor{LightBlue}\textbf{80.2} & \cellcolor{LightBlue}\textbf{51.4} & \cellcolor{LightBlue}\textbf{101.0} & \cellcolor{LightBlue}\textbf{62.5} \\
			\midrule
			\multirow{2}{*}{Replacing Image Encoder Only} &FARE \cite{schlarmann2024robust} & 82.1 & 49.7 & 73.8 & 45.2 & 93.2 & 54.3 \\
			&\cellcolor{LightBlue}\textbf{Ours} & \cellcolor{LightBlue}\textbf{85.9} & \cellcolor{LightBlue}\textbf{55.4} & \cellcolor{LightBlue}\textbf{78.6} & \cellcolor{LightBlue}\textbf{49.8} & \cellcolor{LightBlue}\textbf{98.3} & \cellcolor{LightBlue}\textbf{58.9} \\
			\bottomrule
		\end{tabular}
	}
	\label{tab:FT_VS_Replace}
	\vspace{-0.3cm}
\end{table}

\vspace{0.1cm}
\noindent\textbf{\jh{Fine-tuning target VLM \vs replacing image encoder.}} 
We primarily fine-tune the target  VLM (\eg, BLIP) instead of replacing the image encoder as in FARE \cite{schlarmann2024robust}. Table \ref{tab:FT_VS_Replace}  compares such two settings. 
Our method enjoys better natural performance/adversarial robustness in all cases. 



\begin{table}[!t]
	\vspace{-0.1cm}
	\centering
	\renewcommand{\arraystretch}{1}
	\caption{\jh{Accuracy (\%) of image-text adversaries in fine-tuning.}}
	\vspace{-0.3cm}
	\resizebox{1\linewidth}{!}{
		\begin{tabular}{cccccccc}
			\toprule
			{Adversary Generation} & \multirow{2}{*}{Clean} & \multicolumn{2}{c}{Image Attack} & \multicolumn{2}{c}{Text Attack} & \multicolumn{2}{c}{VLM Attack}\\
			\cmidrule(r){3-4} \cmidrule(r){5-6} \cmidrule(r){7-8}
			During Fine-Tuning&&PGD&AA&BERT-Attack&GBDA&CoAttack&SGA\\
			\midrule
			Image Level Only & 61.27 & 43.42 & 41.83 & 40.53 & 40.28 & 30.85 & 30.19 \\
			\rowcolor{LightBlue}\textbf{Image and Text Levels} & \textbf{62.14} & \textbf{44.34} & \textbf{42.69} & \textbf{41.78} & \textbf{41.55} & \textbf{31.72} & \textbf{31.13} \\
			\bottomrule
		\end{tabular}
	}
	\label{tab:IT_Adv}
\end{table}

\vspace{0.1cm}
\noindent\textbf{Image-text \vs image-only adversaries.} 
As VLM is based on image-text alignment, joint image-text adversaries carry higher boundary risk \cite{zhang2019theoretically} than image- or text-only adversaries \cite{MaoGYWV23,zhang2024adversarial}  so they can robustify the model better. The image-level adv. generation uses PGD, while the text-level generation relies on tokenized embedding to avoid discrete optimization. We use the same adv. generation scheme with other methods during testing.  Table \ref{tab:IT_Adv} shows fine-tuning on image-level adversaries only also improves  robustness.

\begin{table}[!t]
	\vspace{-0.2cm}
	\centering
	\renewcommand{\arraystretch}{1}
	\caption{(Auto-Attack) robust accuracy (\%) \wrt different settings. We also report the robust accuracy against adversarial images \textit{transferred} between base class and superclass classification.}
	\vspace{-0.3cm}
	\resizebox{\linewidth}{!}{
		\begin{tabular}{ccccccc}
			\toprule
			\multirow{2}{*}{Type} & \multicolumn{3}{c}{Base Class} & \multicolumn{3}{c}{Superclass} \\
			\cmidrule(lr){2-4} \cmidrule(lr){5-7}
			& Clean & Robust & \textit{Transfer} & Clean & Robust & \textit{Transfer} \\
			\midrule
			Baseline (TeCoA) & 52.62 & 37.62 & 46.58 & 61.80 & 47.20 & 55.27 \\
			\midrule
			\makecell{Superclass Classification \\ at each level (Euclidean)}$\!\!$ & 53.24 & 38.16 & 48.09 & 68.31 & 55.38 & 63.94 \\
			\rowcolor{LightBlue}\textbf{\makecell{Hyperbolic Space \\ Modeling}} & \textbf{62.14} & \textbf{42.72} & \textbf{63.34} & \textbf{71.68} & \textbf{57.13} & \textbf{66.40} \\
			
			\bottomrule
		\end{tabular}
	}
	\label{tab:15}
	\vspace{-0.3cm}
\end{table}

\vspace{0.1cm}
\noindent\textbf{Hyperbolic \vs Euclidean model.}
Next we analyze robust accuracy in superclass classification. We assess adversarial transferability between superclass and base-class. We also introduce superclass classification in the Euclidean space by averaging feature vectors: $(\mathbf{v}_1, \cdots, \mathbf{v}_N)\rightarrow \frac{1}{N}\sum_{i}\mathbf{v}_i$. Table \ref{tab:15} shows that standard adversarial fine-tuning focused on base categories (Baseline) is low in superclass classification. 
While the Euclidean superclass classification enhances superclass robustness, it does not enhance base category robustness. In contrast, hierarchical modeling  improves robustness in both base and superclasses. 

\vspace{0.1cm}
\noindent\textbf{Hyper-parameter study} is in Appendix \ref{supp:hyperparas}. \textbf{Weighting mechanisms} are studied in Appendix \ref{app:weighting}. \textbf{Base class  representations}  are studied in Appendix \ref{app:averaging}. \textbf{Robustness against black-box multimodal adversaries} is in Appendix \ref{app:black-box}. 

\section{Conclusions}
Motivated by our analysis of adversarial samples targeting superclasses, we have shown that the overemphasis on the base categories lead to poorer adv. robustness in zero-shot settings. Thus, we have leveraged hyperbolic geometry to capture class hierarchies.
By making norms of parent embeddings lower than norms of children embeddings, we  maintain the hierarchical integrity of embedding space. These norms directly impact the margin sizes across hierarchical levels, making each adversarial perturbation carry general-to-specific attack capability to boost robustification.



\section*{Acknowledgments}
This research/project is supported in part by the National Research Foundation, Singapore under its AI Singapore Programme (AISG Award No: AISG3-RP-2022-031), partly by the MTI under its AI Centre of Excellence for Manufacturing (AIMfg) (Award W25MCMF014), the College of Computing and Data Science, Nanyang Technological University (NTU), and the A*STAR-NTU Smart and Sustainable Advanced Manufacturing Research Joint Laboratory. This work was also supported by the CSIRO Allocation Scheme (Lead CI: Piotr Koniusz) and the UNSW Merit Allocation Scheme (Lead CI: Piotr Koniusz) for NCI.

{
    \small
    \bibliographystyle{ieeenat_fullname}
    \bibliography{egbib}

\begin{thebibliography}{56}
\providecommand{\natexlab}[1]{#1}
\providecommand{\url}[1]{\texttt{#1}}
\expandafter\ifx\csname urlstyle\endcsname\relax
  \providecommand{\doi}[1]{doi: #1}\else
  \providecommand{\doi}{doi: \begingroup \urlstyle{rm}\Url}\fi

\bibitem[Bergstra et~al.(2013)Bergstra, Yamins, and
  Cox]{supp_bergstra2013making}
James Bergstra, Daniel Yamins, and David Cox.
\newblock Making a science of model search: Hyperparameter optimization in
  hundreds of dimensions for vision architectures.
\newblock In \emph{International conference on machine learning}, pages
  115--123. PMLR, 2013.

\bibitem[Bossard et~al.(2014)Bossard, Guillaumin, and
  Van~Gool]{supp_bossard2014food}
Lukas Bossard, Matthieu Guillaumin, and Luc Van~Gool.
\newblock Food-101--mining discriminative components with random forests.
\newblock In \emph{Computer Vision--ECCV 2014: 13th European Conference,
  Zurich, Switzerland, September 6-12, 2014, Proceedings, Part VI 13}, pages
  446--461. Springer, 2014.

\bibitem[Bustos et~al.(2020)Bustos, Pertusa, Salinas, and De~La
  Iglesia-Vaya]{supp_bustos2020padchest}
Aurelia Bustos, Antonio Pertusa, Jose-Maria Salinas, and Maria De~La
  Iglesia-Vaya.
\newblock Padchest: A large chest x-ray image dataset with multi-label
  annotated reports.
\newblock \emph{Medical image analysis}, 66:\penalty0 101797, 2020.

\bibitem[Cannon et~al.(1997)Cannon, Floyd, Kenyon, Parry,
  et~al.]{supp_cannon1997hyperbolic}
James~W Cannon, William~J Floyd, Richard Kenyon, Walter~R Parry, et~al.
\newblock Hyperbolic geometry.
\newblock \emph{Flavors of geometry}, 31\penalty0 (59-115):\penalty0 2, 1997.

\bibitem[Carlini and Wagner(2017)]{supp_carlini2017towards}
Nicholas Carlini and David Wagner.
\newblock Towards evaluating the robustness of neural networks.
\newblock In \emph{2017 ieee symposium on security and privacy (sp)}, pages
  39--57. IEEE, 2017.

\bibitem[Cimpoi et~al.(2014)Cimpoi, Maji, Kokkinos, Mohamed, and
  Vedaldi]{supp_cimpoi2014describing}
Mircea Cimpoi, Subhransu Maji, Iasonas Kokkinos, Sammy Mohamed, and Andrea
  Vedaldi.
\newblock Describing textures in the wild.
\newblock In \emph{Proceedings of the IEEE conference on computer vision and
  pattern recognition}, pages 3606--3613, 2014.

\bibitem[Coates et~al.(2011)Coates, Ng, and Lee]{supp_coates2011analysis}
Adam Coates, Andrew Ng, and Honglak Lee.
\newblock An analysis of single-layer networks in unsupervised feature
  learning.
\newblock In \emph{Proceedings of the fourteenth international conference on
  artificial intelligence and statistics}, pages 215--223. JMLR Workshop and
  Conference Proceedings, 2011.

\bibitem[Croce and Hein(2020)]{supp_croce2020reliable}
Francesco Croce and Matthias Hein.
\newblock Reliable evaluation of adversarial robustness with an ensemble of
  diverse parameter-free attacks.
\newblock In \emph{International conference on machine learning}, pages
  2206--2216. PMLR, 2020.

\bibitem[Deng et~al.(2009)Deng, Dong, Socher, Li, Li, and
  Fei-Fei]{supp_deng2009imagenet}
Jia Deng, Wei Dong, Richard Socher, Li-Jia Li, Kai Li, and Li Fei-Fei.
\newblock Imagenet: A large-scale hierarchical image database.
\newblock In \emph{2009 IEEE conference on computer vision and pattern
  recognition}, pages 248--255. Ieee, 2009.

\bibitem[Desai et~al.(2023)Desai, Nickel, Rajpurohit, Johnson, and
  Vedantam]{supp_desai2023hyperbolic}
Karan Desai, Maximilian Nickel, Tanmay Rajpurohit, Justin Johnson, and
  Shanmukha~Ramakrishna Vedantam.
\newblock Hyperbolic image-text representations.
\newblock In \emph{International Conference on Machine Learning}, pages
  7694--7731. PMLR, 2023.

\bibitem[Dosovitskiy et~al.(2021)Dosovitskiy, Beyer, Kolesnikov, Weissenborn,
  Zhai, Unterthiner, Dehghani, Minderer, Heigold, Gelly, Uszkoreit, and
  Houlsby]{supp_DosovitskiyB0WZ21}
Alexey Dosovitskiy, Lucas Beyer, Alexander Kolesnikov, Dirk Weissenborn,
  Xiaohua Zhai, Thomas Unterthiner, Mostafa Dehghani, Matthias Minderer, Georg
  Heigold, Sylvain Gelly, Jakob Uszkoreit, and Neil Houlsby.
\newblock An image is worth 16x16 words: Transformers for image recognition at
  scale.
\newblock In \emph{International Conference on Learning Representations,
  {ICLR}}, 2021.

\bibitem[Fei-Fei et~al.(2004)Fei-Fei, Fergus, and Perona]{supp_fei2004learning}
Li Fei-Fei, Rob Fergus, and Pietro Perona.
\newblock Learning generative visual models from few training examples: An
  incremental bayesian approach tested on 101 object categories.
\newblock In \emph{2004 conference on computer vision and pattern recognition
  workshop}, pages 178--178. IEEE, 2004.

\bibitem[Griffin et~al.(2007)Griffin, Holub, and
  Perona]{supp_griffin2007caltech}
Gregory Griffin, Alex Holub, and Pietro Perona.
\newblock Caltech-256 object category dataset.
\newblock 2007.

\bibitem[Guo et~al.(2021)Guo, Sablayrolles, J{\'{e}}gou, and
  Kiela]{supp_GuoSJK21}
Chuan Guo, Alexandre Sablayrolles, Herv{\'{e}} J{\'{e}}gou, and Douwe Kiela.
\newblock Gradient-based adversarial attacks against text transformers.
\newblock In \emph{Proceedings of the 2021 Conference on Empirical Methods in
  Natural Language Processing, {EMNLP}}, pages 5747--5757, 2021.

\bibitem[Hariharan et~al.(2015)Hariharan, Arbel{\'a}ez, Girshick, and
  Malik]{supp_hariharan2015hypercolumns}
Bharath Hariharan, Pablo Arbel{\'a}ez, Ross Girshick, and Jitendra Malik.
\newblock Hypercolumns for object segmentation and fine-grained localization.
\newblock In \emph{Proceedings of the IEEE conference on computer vision and
  pattern recognition}, pages 447--456, 2015.

\bibitem[Helber et~al.(2019)Helber, Bischke, Dengel, and
  Borth]{supp_helber2019eurosat}
Patrick Helber, Benjamin Bischke, Andreas Dengel, and Damian Borth.
\newblock Eurosat: A novel dataset and deep learning benchmark for land use and
  land cover classification.
\newblock \emph{IEEE Journal of Selected Topics in Applied Earth Observations
  and Remote Sensing}, 12\penalty0 (7):\penalty0 2217--2226, 2019.

\bibitem[Hou et~al.(2020)Hou, Suominen, Koniusz, Caldwell, and
  Gedeon]{supp_Hou2020ATokenwiseC}
Weiwei Hou, Hanna Suominen, Piotr Koniusz, Sabrina Caldwell, and Tom Gedeon.
\newblock A token-wise cnn-based method for sentence compression.
\newblock In \emph{International Conference on Neural Information Processing
  (ICONIP)}, pages 668--679. Springer, Cham, 2020.

\bibitem[Inkawhich et~al.(2023)Inkawhich, McDonald, and
  Luley]{supp_inkawhich2023adversarial}
Nathan Inkawhich, Gwendolyn McDonald, and Ryan Luley.
\newblock Adversarial attacks on foundational vision models.
\newblock \emph{arXiv preprint arXiv:2308.14597}, 2023.

\bibitem[Irvin et~al.(2019)Irvin, Rajpurkar, Ko, Yu, Ciurea-Ilcus, Chute,
  Marklund, Haghgoo, Ball, Shpanskaya, et~al.]{supp_irvin2019chexpert}
Jeremy Irvin, Pranav Rajpurkar, Michael Ko, Yifan Yu, Silviana Ciurea-Ilcus,
  Chris Chute, Henrik Marklund, Behzad Haghgoo, Robyn Ball, Katie Shpanskaya,
  et~al.
\newblock Chexpert: A large chest radiograph dataset with uncertainty labels
  and expert comparison.
\newblock In \emph{Proceedings of the AAAI conference on artificial
  intelligence}, pages 590--597, 2019.

\bibitem[Jia et~al.(2022)Jia, Tang, Chen, Cardie, Belongie, Hariharan, and
  Lim]{supp_jia2022visual}
Menglin Jia, Luming Tang, Bor-Chun Chen, Claire Cardie, Serge Belongie, Bharath
  Hariharan, and Ser-Nam Lim.
\newblock Visual prompt tuning.
\newblock In \emph{European Conference on Computer Vision}, pages 709--727.
  Springer, 2022.

\bibitem[Johnson et~al.(2019)Johnson, Pollard, Berkowitz, Greenbaum, Lungren,
  Deng, Mark, and Horng]{supp_johnson2019mimic}
Alistair~EW Johnson, Tom~J Pollard, Seth~J Berkowitz, Nathaniel~R Greenbaum,
  Matthew~P Lungren, Chih-ying Deng, Roger~G Mark, and Steven Horng.
\newblock Mimic-cxr, a de-identified publicly available database of chest
  radiographs with free-text reports.
\newblock \emph{Scientific data}, 6\penalty0 (1):\penalty0 317, 2019.

\bibitem[Khrulkov et~al.(2020)Khrulkov, Mirvakhabova, Ustinova, Oseledets, and
  Lempitsky]{supp_khrulkov2020hyperbolic}
Valentin Khrulkov, Leyla Mirvakhabova, Evgeniya Ustinova, Ivan Oseledets, and
  Victor Lempitsky.
\newblock Hyperbolic image embeddings.
\newblock In \emph{Proceedings of the IEEE/CVF conference on computer vision
  and pattern recognition}, pages 6418--6428, 2020.

\bibitem[Krause et~al.(2013)Krause, Stark, Deng, and
  Fei-Fei]{supp_krause20133d}
Jonathan Krause, Michael Stark, Jia Deng, and Li Fei-Fei.
\newblock 3d object representations for fine-grained categorization.
\newblock In \emph{Proceedings of the IEEE international conference on computer
  vision workshops}, pages 554--561, 2013.

\bibitem[Krizhevsky et~al.(2009)Krizhevsky, Hinton,
  et~al.]{supp_krizhevsky2009learning}
Alex Krizhevsky, Geoffrey Hinton, et~al.
\newblock Learning multiple layers of features from tiny images.
\newblock 2009.

\bibitem[Lai et~al.(2024)Lai, Yao, Jiang, Wang, He, Tao, and
  Zhou]{supp_lai2024carzero}
Haoran Lai, Qingsong Yao, Zihang Jiang, Rongsheng Wang, Zhiyang He, Xiaodong
  Tao, and S~Kevin Zhou.
\newblock Carzero: Cross-attention alignment for radiology zero-shot
  classification.
\newblock In \emph{Proceedings of the IEEE/CVF Conference on Computer Vision
  and Pattern Recognition}, pages 11137--11146, 2024.

\bibitem[Lee et~al.(2020)Lee, Yoon, Kim, Kim, Kim, So, and
  Kang]{supp_lee2020biobert}
Jinhyuk Lee, Wonjin Yoon, Sungdong Kim, Donghyeon Kim, Sunkyu Kim, Chan~Ho So,
  and Jaewoo Kang.
\newblock Biobert: a pre-trained biomedical language representation model for
  biomedical text mining.
\newblock \emph{Bioinformatics}, 36\penalty0 (4):\penalty0 1234--1240, 2020.

\bibitem[Li et~al.(2022)Li, Li, Xiong, and Hoi]{supp_li2022blip}
Junnan Li, Dongxu Li, Caiming Xiong, and Steven Hoi.
\newblock Blip: Bootstrapping language-image pre-training for unified
  vision-language understanding and generation.
\newblock In \emph{ICML}, 2022.

\bibitem[Li et~al.(2020)Li, Ma, Guo, Xue, and Qiu]{supp_LiMGXQ20}
Linyang Li, Ruotian Ma, Qipeng Guo, Xiangyang Xue, and Xipeng Qiu.
\newblock {BERT-ATTACK:} adversarial attack against {BERT} using {BERT}.
\newblock In \emph{Proceedings of the 2020 Conference on Empirical Methods in
  Natural Language Processing {EMNLP}}, pages 6193--6202, 2020.

\bibitem[Li et~al.(2024)Li, Guan, Qiu, and Spratling]{supp_li2024one}
Lin Li, Haoyan Guan, Jianing Qiu, and Michael Spratling.
\newblock One prompt word is enough to boost adversarial robustness for
  pre-trained vision-language models.
\newblock In \emph{Proceedings of the IEEE/CVF conference on computer vision
  and pattern recognition}, 2024.

\bibitem[Lin et~al.(2017)Lin, Doll{\'a}r, Girshick, He, Hariharan, and
  Belongie]{supp_lin2017feature}
Tsung-Yi Lin, Piotr Doll{\'a}r, Ross Girshick, Kaiming He, Bharath Hariharan,
  and Serge Belongie.
\newblock Feature pyramid networks for object detection.
\newblock In \emph{Proceedings of the IEEE conference on computer vision and
  pattern recognition}, pages 2117--2125, 2017.

\bibitem[Lu et~al.(2025)Lu, Liu, and Koniusz]{supp_tokenize_keypoints}
Changsheng Lu, Zheyuan Liu, and Piotr Koniusz.
\newblock Openkd: Opening prompt diversity for zero- and few-shot keypoint
  detection.
\newblock In \emph{Computer Vision -- ECCV 2024}, pages 148--165, Cham, 2025.
  Springer Nature Switzerland.

\bibitem[Lu et~al.(2023)Lu, Wang, Wang, Guan, Gao, and Zheng]{supp_lu2023set}
Dong Lu, Zhiqiang Wang, Teng Wang, Weili Guan, Hongchang Gao, and Feng Zheng.
\newblock Set-level guidance attack: Boosting adversarial transferability of
  vision-language pre-training models.
\newblock In \emph{Proceedings of the IEEE/CVF International Conference on
  Computer Vision}, pages 102--111, 2023.

\bibitem[Madry et~al.(2018)Madry, Makelov, Schmidt, Tsipras, and
  Vladu]{supp_MadryMSTV18}
Aleksander Madry, Aleksandar Makelov, Ludwig Schmidt, Dimitris Tsipras, and
  Adrian Vladu.
\newblock Towards deep learning models resistant to adversarial attacks.
\newblock In \emph{6th International Conference on Learning Representations,
  {ICLR}}, 2018.

\bibitem[Maji et~al.(2013)Maji, Rahtu, Kannala, Blaschko, and
  Vedaldi]{supp_maji2013fine}
Subhransu Maji, Esa Rahtu, Juho Kannala, Matthew Blaschko, and Andrea Vedaldi.
\newblock Fine-grained visual classification of aircraft.
\newblock \emph{arXiv preprint arXiv:1306.5151}, 2013.

\bibitem[Mao et~al.(2023)Mao, Geng, Yang, Wang, and Vondrick]{supp_MaoGYWV23}
Chengzhi Mao, Scott Geng, Junfeng Yang, Xin Wang, and Carl Vondrick.
\newblock Understanding zero-shot adversarial robustness for large-scale
  models.
\newblock In \emph{The Eleventh International Conference on Learning
  Representations,{ICLR}}, 2023.

\bibitem[Miller(1995)]{supp_miller1995wordnet}
George~A Miller.
\newblock Wordnet: a lexical database for english.
\newblock \emph{Communications of the ACM}, 38\penalty0 (11):\penalty0 39--41,
  1995.

\bibitem[Ni et~al.(2024)Ni, Zhang, and Koniusz]{supp_ni2024pace}
Yao Ni, Shan Zhang, and Piotr Koniusz.
\newblock {PACE}: marrying the generalization of {PA}rameter-efficient
  fine-tuning with consistency regularization.
\newblock In \emph{The Thirty-eighth Annual Conference on Neural Information
  Processing Systems}, 2024.

\bibitem[Nilsback and Zisserman(2008)]{supp_nilsback2008automated}
Maria-Elena Nilsback and Andrew Zisserman.
\newblock Automated flower classification over a large number of classes.
\newblock In \emph{2008 Sixth Indian conference on computer vision, graphics \&
  image processing}, pages 722--729. IEEE, 2008.

\bibitem[OpenAI(2024)]{supp_openai_chatgpt_2024}
OpenAI.
\newblock Chatgpt [large language model].
\newblock \url{https://chatgpt.com}, 2024.

\bibitem[Parkhi et~al.(2012)Parkhi, Vedaldi, Zisserman, and
  Jawahar]{supp_parkhi2012cats}
Omkar~M Parkhi, Andrea Vedaldi, Andrew Zisserman, and CV Jawahar.
\newblock Cats and dogs.
\newblock In \emph{2012 IEEE conference on computer vision and pattern
  recognition}, pages 3498--3505. IEEE, 2012.

\bibitem[Pellegrini et~al.(2023)Pellegrini, Keicher, {\"O}zsoy, Jiraskova,
  Braren, and Navab]{supp_pellegrini2023xplainer}
Chantal Pellegrini, Matthias Keicher, Ege {\"O}zsoy, Petra Jiraskova, Rickmer
  Braren, and Nassir Navab.
\newblock Xplainer: From x-ray observations to explainable zero-shot diagnosis.
\newblock In \emph{International Conference on Medical Image Computing and
  Computer-Assisted Intervention}, pages 420--429. Springer, 2023.

\bibitem[Radford et~al.(2021)Radford, Kim, Hallacy, Ramesh, Goh, Agarwal,
  Sastry, Askell, Mishkin, Clark, et~al.]{supp_radford2021learning}
Alec Radford, Jong~Wook Kim, Chris Hallacy, Aditya Ramesh, Gabriel Goh,
  Sandhini Agarwal, Girish Sastry, Amanda Askell, Pamela Mishkin, Jack Clark,
  et~al.
\newblock Learning transferable visual models from natural language
  supervision.
\newblock In \emph{International conference on machine learning}, pages
  8748--8763. PMLR, 2021.

\bibitem[Schlarmann et~al.(2024)Schlarmann, Singh, Croce, and
  Hein]{supp_schlarmann2024robust}
Christian Schlarmann, Naman~Deep Singh, Francesco Croce, and Matthias Hein.
\newblock Robust clip: Unsupervised adversarial fine-tuning of vision
  embeddings for robust large vision-language models.
\newblock \emph{arXiv preprint arXiv:2402.12336}, 2024.

\bibitem[Tiu et~al.(2022)Tiu, Talius, Patel, Langlotz, Ng, and
  Rajpurkar]{supp_tiu2022expert}
Ekin Tiu, Ellie Talius, Pujan Patel, Curtis~P Langlotz, Andrew~Y Ng, and Pranav
  Rajpurkar.
\newblock Expert-level detection of pathologies from unannotated chest x-ray
  images via self-supervised learning.
\newblock \emph{Nature Biomedical Engineering}, 6\penalty0 (12):\penalty0
  1399--1406, 2022.

\bibitem[Veeling et~al.(2018)Veeling, Linmans, Winkens, Cohen, and
  Welling]{supp_veeling2018rotation}
Bastiaan~S Veeling, Jasper Linmans, Jim Winkens, Taco Cohen, and Max Welling.
\newblock Rotation equivariant cnns for digital pathology.
\newblock In \emph{Medical Image Computing and Computer Assisted
  Intervention--MICCAI 2018: 21st International Conference, Granada, Spain,
  September 16-20, 2018, Proceedings, Part II 11}, pages 210--218. Springer,
  2018.

\bibitem[Wang et~al.(2024)Wang, Zhang, Yuan, and Shan]{supp_wang2024pre}
Sibo Wang, Jie Zhang, Zheng Yuan, and Shiguang Shan.
\newblock Pre-trained model guided fine-tuning for zero-shot adversarial
  robustness.
\newblock In \emph{Proceedings of the IEEE/CVF conference on computer vision
  and pattern recognition}, 2024.

\bibitem[Wang et~al.(2017)Wang, Peng, Lu, Lu, Bagheri, and
  Summers]{supp_wang2017chestx}
Xiaosong Wang, Yifan Peng, Le Lu, Zhiyong Lu, Mohammadhadi Bagheri, and
  Ronald~M Summers.
\newblock Chestx-ray8: Hospital-scale chest x-ray database and benchmarks on
  weakly-supervised classification and localization of common thorax diseases.
\newblock In \emph{Proceedings of the IEEE conference on computer vision and
  pattern recognition}, pages 2097--2106, 2017.

\bibitem[Xiao et~al.(2010)Xiao, Hays, Ehinger, Oliva, and
  Torralba]{supp_xiao2010sun}
Jianxiong Xiao, James Hays, Krista~A Ehinger, Aude Oliva, and Antonio Torralba.
\newblock Sun database: Large-scale scene recognition from abbey to zoo.
\newblock In \emph{2010 IEEE computer society conference on computer vision and
  pattern recognition}, pages 3485--3492. IEEE, 2010.

\bibitem[Yu et~al.(2022)Yu, Wang, Vasudevan, Yeung, Seyedhosseini, and
  Wu]{supp_YuWVYSW22}
Jiahui Yu, Zirui Wang, Vijay Vasudevan, Legg Yeung, Mojtaba Seyedhosseini, and
  Yonghui Wu.
\newblock Coca: Contrastive captioners are image-text foundation models.
\newblock \emph{Trans. Mach. Learn. Res.}, 2022, 2022.

\bibitem[Zhang et~al.(2022)Zhang, Yi, and Sang]{supp_zhang2022towards}
Jiaming Zhang, Qi Yi, and Jitao Sang.
\newblock Towards adversarial attack on vision-language pre-training models.
\newblock In \emph{Proceedings of the 30th ACM International Conference on
  Multimedia}, pages 5005--5013, 2022.

\bibitem[Zhang et~al.(2024)Zhang, Zhu, Liu, Yu, Koniusz, and
  King]{supp_zhang2024moreextremegradientboost}
Yifei Zhang, Hao Zhu, Aiwei Liu, Han Yu, Piotr Koniusz, and Irwin King.
\newblock Less is more: Extreme gradient boost rank-1 adaption for efficient
  finetuning of llms.
\newblock In \emph{arXiv/2410.19694}, 2024.

\bibitem[Zhang et~al.(2025)Zhang, Zhu, Dong, Shi, Meng, Koniusz, and
  Yu]{supp_zhang2025crossspectra}
Yifei Zhang, Hao Zhu, Junhao Dong, Haoran Shi, Ziqiao Meng, Piotr Koniusz, and
  Han Yu.
\newblock Crossspectra: Exploiting cross-layer smoothness for
  parameter-efficient fine-tuning.
\newblock In \emph{The Thirty-ninth Annual Conference on Neural Information
  Processing Systems}, 2025.

\bibitem[Zhao et~al.(2023{\natexlab{a}})Zhao, Pang, Du, Yang, Li, Cheung, and
  Lin]{supp_zhao2023evaluating}
Yunqing Zhao, Tianyu Pang, Chao Du, Xiao Yang, Chongxuan Li, Ngai-Man~Man
  Cheung, and Min Lin.
\newblock On evaluating adversarial robustness of large vision-language models.
\newblock \emph{Advances in Neural Information Processing Systems},
  36:\penalty0 54111--54138, 2023{\natexlab{a}}.

\bibitem[Zhao et~al.(2023{\natexlab{b}})Zhao, Liu, Wu, Li, Wang, Teng, Liu, Li,
  Cui, Wang, et~al.]{supp_zhao2023clip}
Zihao Zhao, Yuxiao Liu, Han Wu, Yonghao Li, Sheng Wang, Lin Teng, Disheng Liu,
  Xiang Li, Zhiming Cui, Qian Wang, et~al.
\newblock Clip in medical imaging: A comprehensive survey.
\newblock \emph{arXiv preprint arXiv:2312.07353}, 2023{\natexlab{b}}.

\bibitem[Zhou et~al.(2022)Zhou, Yang, Loy, and Liu]{supp_zhou2022coop}
Kaiyang Zhou, Jingkang Yang, Chen~Change Loy, and Ziwei Liu.
\newblock Learning to prompt for vision-language models.
\newblock \emph{International Journal of Computer Vision (IJCV)}, 2022.

\bibitem[Zhu et~al.(2025)Zhu, Zhang, Dong, and Koniusz]{supp_Zhu_2025_CVPR}
Hao Zhu, Yifei Zhang, Junhao Dong, and Piotr Koniusz.
\newblock Bilora: Almost-orthogonal parameter spaces for continual learning.
\newblock In \emph{Proceedings of the IEEE/CVF Conference on Computer Vision
  and Pattern Recognition (CVPR)}, pages 25613--25622, 2025.

\end{thebibliography}
}


\onecolumn
\let\thetitle\relax

\appendix

\title{Hierarchically Robust Zero-shot Vision-Language Models\\ -- \textit{Supplementary Material} -- }

\maketitle

\begingroup
\renewcommand\thefootnote{}
\footnotetext{ \hspace{-3ex} \Letter\ Corresponding authors.}
\endgroup

\hypersetup{citecolor=white}

\hypersetup{citecolor=ourblue}

\setcounter{table}{16}
\setcounter{equation}{16}
\setcounter{figure}{3}

\begin{abstract}
In this supplementary material, we commence with a comprehensive discussion of the pipeline of our proposed adversarial fine-tuning method based on our hyperbolic mechanism in Appendix \ref{supp:pipe}. Moreover, we elaborate on our experimental configurations, comprising both the dataset description and the implementation details (including the extension in the context of BLIP and Medical CLIP) in Appendix \ref{supp:exp_details}. We also provide further details regarding our category-based hierarchy construction mechanism in Appendix \ref{supp:Superclass}. Hyper-parameter analyses are presented in Appendix \ref{supp:hyperparas}. We then discuss the impact of diverse intra-class variability strategies (Eq. (\ref{eq:8}\&\ref{eq:9})) in Appendix \ref{supp:intra_variability}.
\end{abstract}

\begin{figure}[!ht]
\centering
\includegraphics[width=1\linewidth]{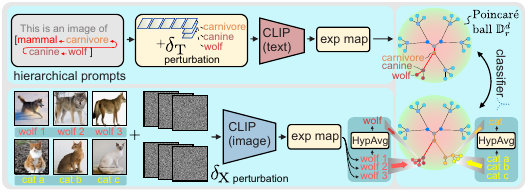}
\vspace{-0.4cm}
\caption{Our pipeline.}
\label{supp_fig:pipe}
\vspace{-0.4cm}
\end{figure}

\section{Pipeline}
\label{supp:pipe}

Figure \ref{supp_fig:pipe} is a zoom of our pipeline in Figure \ref{fig:pipe}. Two branches of CLIP are used, \ie, text encoder and image encoder. For text, for each category, we look up its hierarchy in a predefined hierarchical tree (\eg, for ImageNet, the categories follow WorldNet) and extract the path from the root all the way to the leaf category. For each category level, we form one text prompt and encode with CLIP that firstly tokenizes text \citelatex{supp_schlarmann2024robust,supp_zhao2023clip,supp_Hou2020ATokenwiseC,supp_tokenize_keypoints}. Notice $\boldsymbol{\delta}_\text{T}$ is added to the contextual part of prompts for adversarial text learning. The exponential map elevates embeddings from the Euclidean space into the hyperbolic space (Poincaré ball).

For image, in each mini-batch, we have several images of the same leaf (base) category. We embed them via the image branch of the CLIP. Notice we add $\boldsymbol{\delta}_\text{X}$ per image to learn adversarial perturbations. Subsequently, The exponential map elevates image embeddings from the Euclidean space into the hyperbolic space. Here, in order to obtain embedding representing {\em wolf}, we use Hyperbolic averaging. \ie, $\operatorname{HypAvg}(\cdot)$. The same Hyperbolic averaging strategy is used to obtain further embeddings across hierarchical levels.





\begin{figure*}[!ht]
\centering
\includegraphics[width=0.8\linewidth]{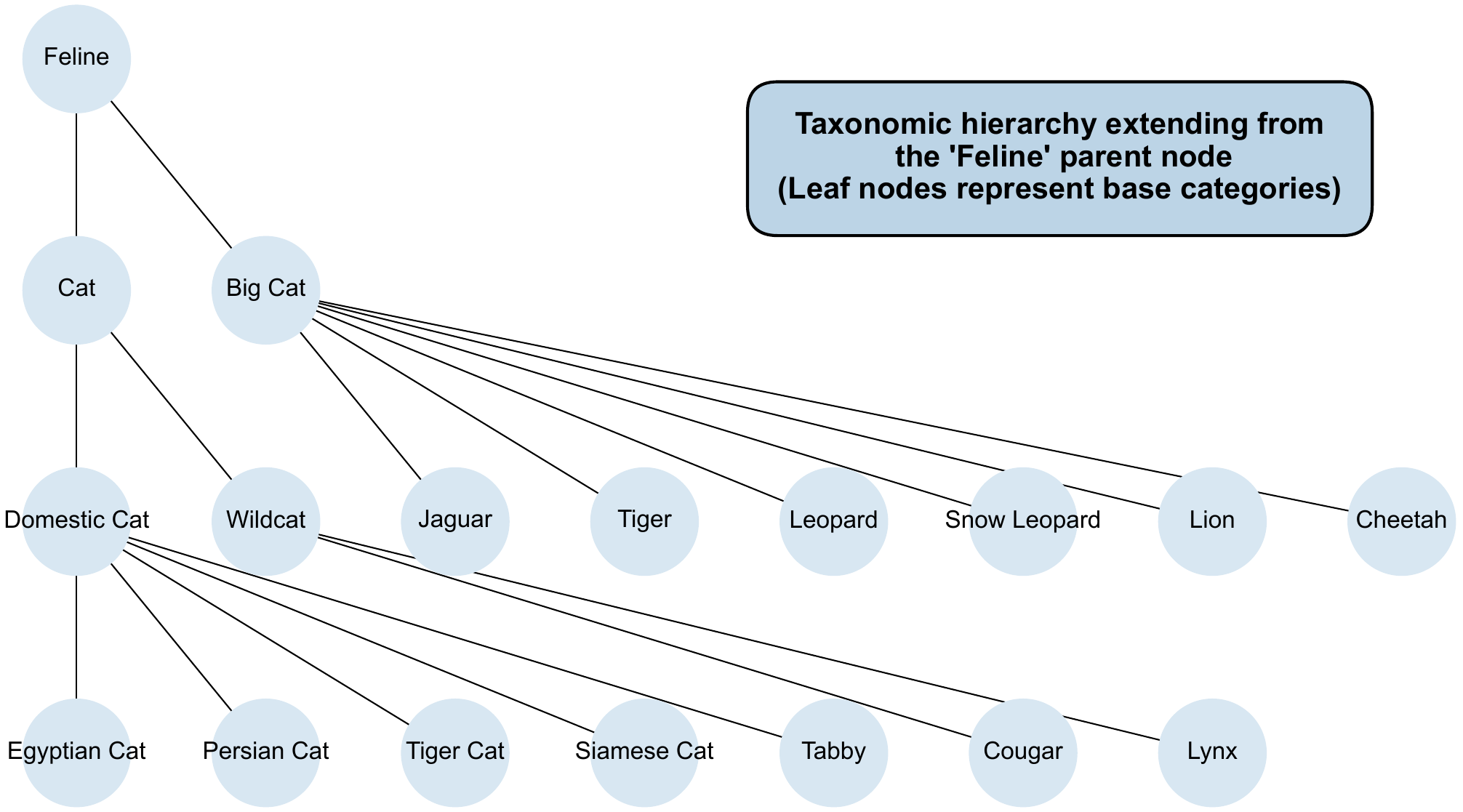}
\vspace{-0.4cm}
\caption{The hierarchy of a sub-branch in ImageNet with the root node of ``Feline''. The leaf nodes represent the original base classes.}
\label{supp_fig:hierarchy}
\vspace{-0.4cm}
\end{figure*}

\begin{figure*}[!ht]
\centering
\includegraphics[width=0.8\linewidth]{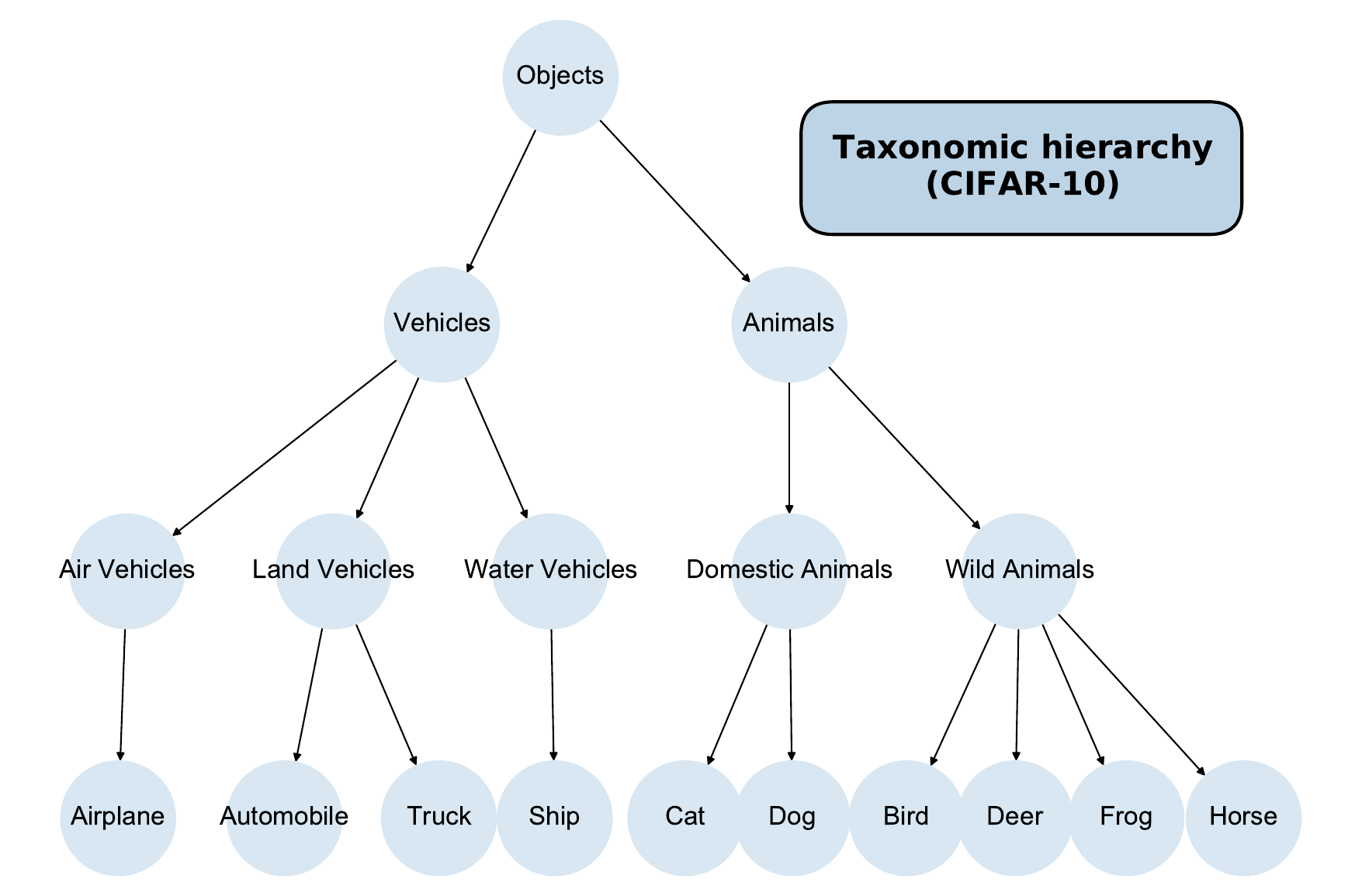}
\vspace{-0.4cm}
\caption{The hierarchy of CIFAR-10. The leaf nodes represent the original base classes.}
\label{supp_fig:hierarchy_cifar}
\end{figure*}

\section{Experimental Details}
\label{supp:exp_details}
This section presents a comprehensive overview of the experimental configurations used in our study, including detailed information about the datasets used for adversarially robust fine-tuning and the corresponding implementation details of our proposed method.

\subsection{Datasets}
\label{supp:dataset_details}
In accordance with the evaluation protocols established in prior research \citelatex{supp_MaoGYWV23, supp_wang2024pre}, we conduct adversarial fine-tuning of the CLIP model using the ImageNet training dataset \citelatex{supp_deng2009imagenet}. To systematically evaluate the robustness of our fine-tuned model, we test it on the ImageNet validation set\textemdash as ImageNet typically provides a validation split used for testing purposes, with a subset of the training data reserved for validation—and on an additional set of 14 zero-shot datasets that span a diverse array of image recognition tasks. Collectively, these 15 datasets are categorized into four groups:
\begin{itemize} 
	\item \textbf{General Image Classification}: ImageNet \citelatex{supp_deng2009imagenet}, STL-10 \citelatex{supp_coates2011analysis}, CIFAR-10 and CIFAR-100 \citelatex{supp_krizhevsky2009learning}, Caltech-101 \citelatex{supp_fei2004learning}, and Caltech-256 \citelatex{supp_griffin2007caltech}. 
	
	\item \textbf{Fine-Grained Classification}: FGVC Aircraft \citelatex{supp_maji2013fine}, Flower102 \citelatex{supp_nilsback2008automated}, Food101 \citelatex{supp_bossard2014food}, Oxford-IIIT Pets \citelatex{supp_parkhi2012cats}, and Stanford Cars \citelatex{supp_krause20133d}. 
	
	\item \textbf{Domain-Specific Classification}: Describable Textures Dataset (DTD) \citelatex{supp_cimpoi2014describing}, EuroSAT \citelatex{supp_helber2019eurosat}, and PatchCamelyon (PCAM) \citelatex{supp_veeling2018rotation}. 
	
	\item \textbf{Scene Recognition}: SUN397 \citelatex{supp_xiao2010sun}. 
\end{itemize}
During adversarial fine-tuning, we apply standardized data preprocessing techniques to ensure consistency across all datasets. Specifically, each input image is resized to a resolution of $224 \times 224$ pixels, followed by a center crop operation. This preprocessing aligns with common practices in fine-tuning vision-language models and facilitates fair comparisons by maintaining uniform input dimensions.

\subsection{Implementation}
\label{supp:implement_details}

\vspace{0.1cm} 
\noindent \textbf{Standard setup.}
In line with the configurations used in prior studies \citelatex{supp_MaoGYWV23, supp_wang2024pre}, we adopt the CLIP model \citelatex{supp_radford2021learning} utilizing the Vision Transformer (ViT) architecture, specifically ViT-Base/32 \citelatex{supp_DosovitskiyB0WZ21}. For network parameter optimization during adversarial fine-tuning, we employ the Stochastic Gradient Descent (SGD) optimizer with a momentum of 0.9 and a batch size of 512. The learning rate is initialized at $1 \times 10^{-5}$ and scheduled using cosine annealing for the fine-tuning of the whole vision encoder and the projection layer of the text encoder. When implementing Visual Prompt Tuning (VPT) \citelatex{supp_jia2022visual}, a parameter-efficient fine-tuning strategy, we introduce token-level learnable parameters of size 100 into the vision branch of CLIP and set the learning rate to 40. During training, adversarial examples are generated at both the image and text levels using Projected Gradient Descent (PGD) \citelatex{supp_MadryMSTV18} with $3$ iterations. For image-level adversarial perturbations, we adopt the $\ell_{\infty}$-norm threat model with a maximum perturbation radius of $\epsilon_{\text{X}} = 1/255$ and a step size of $\alpha_{\text{X}} = 1/255$, unless specified otherwise. For text-level adversarial perturbations\textemdash applied exclusively during fine-tuning\textemdash we set the step size to $\alpha_{\text{T}} = 1 \times 10^{-4}$ and the perturbation radius to $\epsilon_{\text{T}} = 2 \times 10^{-4}$. Superclasses are constructed using the ImageNet hierarchy up to $L=5$ for both image and text modalities. We set the curvature of the hyperbolic space to $r=1.0$. The projection hyper-parameter is set to $\xi=1\times 10^{-5}$ to prevent features from reaching the Poincaré disk boundary. The vicinity radius around text embeddings is configured as $\zeta_\text{vic}=5\times 10^{-2}$, and the margin factor is set to $\zeta_\text{gap}=1\times 10^{-2}$. To balance the contributions of different loss components, we assign the loss weighting factors as $\lambda_1=0.3$ and $\lambda_2=0.1$. Note that the setting of all the hyper-parameters is obtained through the Hyperopt package \citelatex{supp_bergstra2013making} for a $25$-iteration hyper-parameter search on a 1\% subset of the ImageNet training set. The hyper-parameter setting was then applied without tuning to adversarial fine-tuning of all other scenarios. All experiments are conducted on eight NVIDIA Tesla A100 GPUs.

\vspace{0.1cm} 
\noindent \textbf{Evaluation protocol.}
Aligned with previous research on adversarially robust CLIP fine-tuning \citelatex{supp_MaoGYWV23, supp_wang2024pre}, we focus on evaluating robustness against three strong white-box adversarial attacks: 20-step PGD \citelatex{supp_MadryMSTV18}, the Carlini and Wagner (CW) attack \citelatex{supp_carlini2017towards}, and Auto-Attack (AA) \citelatex{supp_croce2020reliable}. In addition to image-level attacks, we also assess robustness against \textit{text-level attacks} such as BERT-Attack \citelatex{supp_LiMGXQ20} and Gradient-Based Distributional Attack (GBDA) \citelatex{supp_GuoSJK21}, as well as \textit{bi-level attacks} using Collaborative Multimodal Adversarial Attack (Co-Attack) \citelatex{supp_zhang2022towards} and Set-level Guidance Attack (SGA) \citelatex{supp_lu2023set}, which are discussed and evaluated in the main text.

\vspace{0.1cm} 
\noindent\textbf{\jh{Experimental settings for BLIP/CoCa extension.}}
\jh{To assess our method's zero-shot robustness on downstream tasks, we expand our experiments to integrating the BLIP architecture \citelatex{supp_li2022blip}, a large-scale vision-language model that unifies vision-language understanding tasks through bootstrapped language-image pre-training. Specifically, we evaluate zero-shot adversarial robustness on two tasks: image-text retrieval and image captioning. Following the pipeline in \citelatex{supp_li2022blip}, we adversarially optimize the Image-Text Contrastive (ITC) loss, Image-Text Matching (ITM) loss, and Language Modeling (LM) loss to obtain a robust version of BLIP. For the CoCa architecture \citelatex{supp_YuWVYSW22}, we adversarially optimize the ITC loss. Subsequently, we assess robustness by applying the iterative PGD attack method, using the ITM loss for image-text retrieval and the LM loss for image captioning. We follow the same perturbation radius ($\epsilon_\text{V} = 1/255$ and $\epsilon_{\text{T}} = 2 \times 10^{-4}$) during fine-tuning as in our original manuscript, with the sole modification being the replacement of the objective function for adversary generation.}

\vspace{0.1cm} 
\noindent\textbf{Experimental settings for medical CLIP extension.}
To expand our empirical analyses for robust medical imaging, we utilize a CLIP model pre-trained specifically on radiology datasets following the CheXzero framework \citelatex{supp_tiu2022expert} with the architecture of ViT-B/16. In line with established protocols \citelatex{supp_tiu2022expert, supp_pellegrini2023xplainer, supp_lai2024carzero}, we utilize the MIMIC dataset \citelatex{supp_johnson2019mimic}\textemdash a systematic database of chest radiographs paired with detailed radiology reports\textemdash for adversarial fine-tuning. The text encoder in our CLIP model leverages BioBERT \citelatex{supp_lee2020biobert}, a specialized biomedical language model optimized for text mining for biomedical analysis. In zero-shot scenarios, we evaluate the robust CLIP models on three widely-used multi-label radiology datasets: ChestX-ray14 \citelatex{supp_wang2017chestx}, CheXpert \citelatex{supp_irvin2019chexpert}, and PadChest \citelatex{supp_bustos2020padchest}. We report the Area Under the Curve (AUC) metrics for both clean and adversarial images. The adversaries are generated based on a 20-step PGD approach with $\epsilon_\text{V} = 1/255$ and $\epsilon_{\text{T}} = 2 \times 10^{-4}$.

\section{Hierarchy Construction}
\label{supp:Superclass}

For the ImageNet dataset \citelatex{supp_deng2009imagenet}, we leverage the hierarchical taxonomy provided by WordNet \citelatex{supp_miller1995wordnet} to construct superclasses by extracting hypernyms of synsets. This hierarchical structure enables us to represent semantic relationships between concepts at multiple levels of abstraction. For instance, Figure \ref{supp_fig:hierarchy} illustrates a sub-branch of the ImageNet hierarchy rooted at the node \textit{Feline}. In this hierarchy, each base category\textemdash such as a specific species of domestic cat\textemdash, is connected to its immediate superclass, recursively forming a tree that ascends to more abstract concepts like \textit{Mammal} and \textit{Animal}. This method allows us to obtain superclasses across diverse hierarchical levels, effectively capturing varying granularities of class concepts and enriching the semantic context for each category.

In scenarios where datasets lack predefined hierarchical taxonomies, we generate superclasses using a Large Language Model (LLM), specifically ChatGPT-4o \citelatex{supp_openai_chatgpt_2024}. We design prompts that instruct the LLM to suggest contextually appropriate superclasses for each base class. For instance, we could query, \texttt{``Provide a general category-based hierarchy that is built upon all the [base classes].''} To ensure the generated superclasses are semantically coherent and align with the dataset's context, we perform manual reviews and validations. This process involves cross-referencing the suggested superclasses with domain knowledge and existing literature to confirm their relevance and accuracy. By systematically applying this approach, we establish consistent hierarchical structures across diverse datasets, which facilitates the implementation of our hyperbolic space modeling for robust image-text alignment. For instance, the hierarchical tree of categories for the CIFAR-10 dataset \citelatex{supp_deng2009imagenet} is illustrated in Figure \ref{supp_fig:hierarchy_cifar}.


\begin{figure}[!t]
	\centering
	\begin{subfigure}[t]{0.25\linewidth} 
		\centering
		\includegraphics[width=1\linewidth]{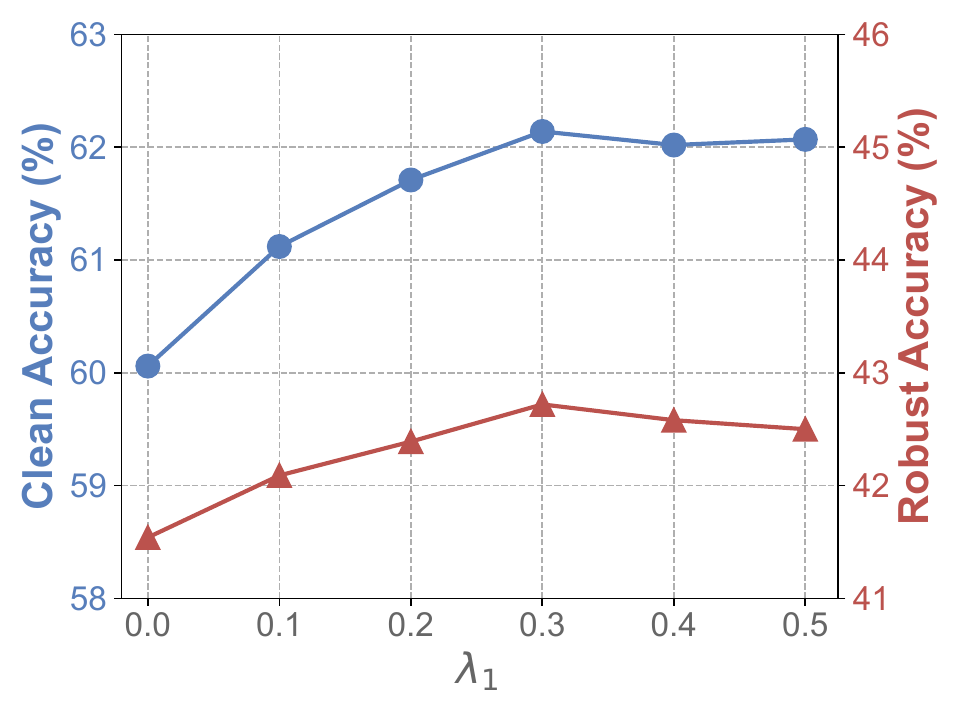}
		\vspace{-0.6cm}
		\caption{}
		\label{fig:2_1}
		\vspace{0.2cm}
	\end{subfigure}
	\hspace{0.5cm}
	\begin{subfigure}[t]{0.25\linewidth} 
		\centering
		\includegraphics[width=1\linewidth]{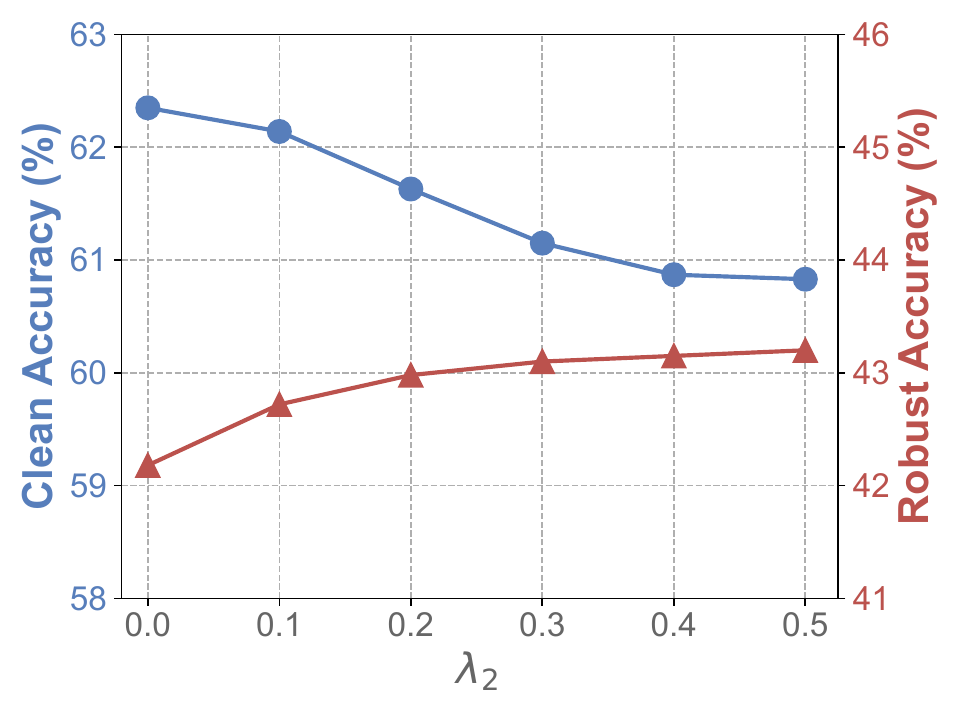}
		\vspace{-0.6cm}
		\caption{}
		\label{fig:2_2}
		\vspace{0.2cm}
	\end{subfigure}
	\vspace{-0.5cm}
	\caption{Hyper-parameter ($\lambda_1$ and $\lambda_2$) sensitivity of our method on average clean and (Auto-Attack) robust accuracy (\%).}
	\label{fig:2}
\end{figure}

\begin{figure}[!t]
        \centering
	\begin{subfigure}[t]{0.25\linewidth} 
		\centering
		\includegraphics[width=1\linewidth]{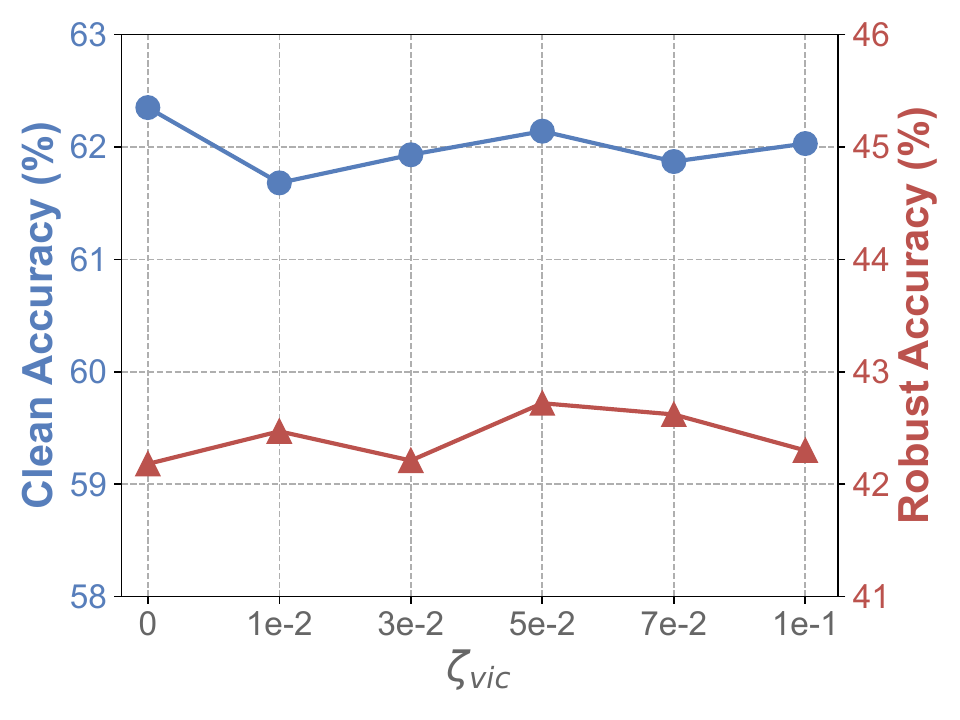}
		\vspace{-0.6cm}
		\caption{}
		\label{supp_fig:hyperpara_1}
		\vspace{0.2cm}
	\end{subfigure}
	\hspace{0.5cm}
	\begin{subfigure}[t]{0.25\linewidth} 
		\centering
		\includegraphics[width=1\linewidth]{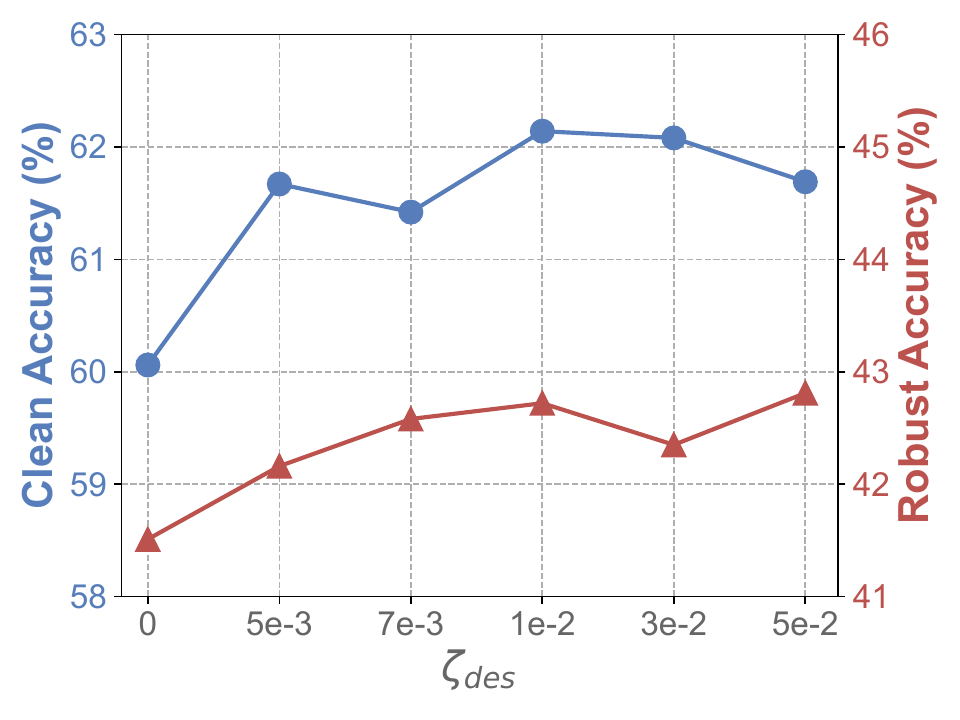}
		\vspace{-0.6cm}
		\caption{}
		\label{supp_fig:hyperpara_2}
		\vspace{0.2cm}
	\end{subfigure}
	\vspace{-0.5cm}
	\caption{Hyper-parameter ($\zeta_\text{vic}$ and $\zeta_\text{gap}$) sensitivity of our adversarial fine-tuning method on average clean and (Auto-Attack) robust accuracy (\%) across 15 datasets in the zero-shot setting.}
	\label{supp_fig:hyperpara}
\end{figure}

\section{Hyper-parameter Analyses}
\label{supp:hyperparas}

For a better understanding of our method, we analyze how the integrated hyperparameters affect performance. Specifically, in Figure \ref{fig:2}, we present the average accuracy on both clean samples and those attacked by AutoAttack across 15 datasets in the zero-shot setting. The results show that increasing the hyperparameter $\lambda_1$, which circumvents the inductive bias towards static alignment reference, leads to an improved trade-off between accuracy and robustness. Meanwhile, enhancing the value of $\lambda_2$ results in improved adversarial robustness but a corresponding decrease in natural performance. This indicates that the weighting factor $\lambda_2$ associated with the norm descent constraints in hyperbolic space regulates the balance between natural accuracy and adversarial robustness.

In addition to the loss weighting factors $\lambda_1$ and $\lambda_2$ we discussed above, we explore the impact of other hyper-parameters in our hierarchical adversarial fine-tuning approach. As shown in Figure \ref{supp_fig:hyperpara}, we report both clean and (Auto-Attack) robust accuracy of our method under diverse hyper-parameter settings. Note that all hyper-parameters in this analysis were tuned based on a tiny subset of the ImageNet training set to ensure fairness. This hyper-parameter setting is then applied to the adversarial fine-tuning of diverse scenarios. We can observe that appropriately choosing the margin factors $\zeta_\text{vic}$ and $\zeta_\text{gap}$ for adversarial fine-tuning can lead to a reasonable trade-off between natural performance and adversarial robustness in the zero-shot setting.

\section{Impact of Intra-class Variability}
\label{supp:intra_variability}
In Section \ref{sec:intra_varia}, we discussed how incorporating intra-class variability can mitigate the inductive bias toward static alignment references, leading to an improved trade-off between accuracy and robustness. Specifically, we align adversarial image embeddings within the upper arc vicinity of the base class text embedding while maintaining the hierarchical order of these embeddings. Extending this concept, we also explore the impact of intra-class variability for clean embeddings, as presented in Table \ref{supp_tab:1}. Our findings indicate that while introducing intra-class variability for clean embeddings can modestly enhance natural performance, it results in a substantial decline in adversarial robustness in the zero-shot setting. Therefore, we choose to apply intra-class variability exclusively to adversarial embeddings for each input image.

\begin{table}[h]
	\centering
	\renewcommand{\arraystretch}{1}
	\caption{Zero-shot clean and robust accuracy (\%) \wrt diverse intra-class variability configurations.}
	\vspace{-0.3cm}
	\resizebox{0.5\linewidth}{!}{
		\begin{tabular}{cccc}
			\toprule
			\makecell{Intra-Class Variability Setting} & Clean & PGD & AA  \\
			\midrule
			Clean \& Adversarial Embeddings & 62.56 & 41.30 & 40.12 \\
			\rowcolor{LightBlue}Adversarial Embeddings Only & \textbf{62.14} & \textbf{44.34} & \textbf{42.72} \\
			
			\bottomrule
		\end{tabular}
	}
	\label{supp_tab:1}
\end{table}

\section{\jh{Impact of Augmenting the Hierarchical Trees to Forests}}
\jh{Recall that we extend our taxonomic hierarchy by constructing multiple hyperbolic trees based on diverse taxonomic structures obtained through LLMs. Specifically, we query LLMs to generate alternative hierarchical taxonomies, capturing different semantic relationships among categories. By incorporating multiple hierarchical trees, we aim to enhance robustness by leveraging richer multi-level abstractions. Table \ref{tab:multi_trees} presents a comparative analysis of model performance with varying numbers of hyperbolic trees. Our findings indicate that increasing the number of trees consistently improves zero-shot adversarial robustness while maintaining competitive clean accuracy. However, we observe diminishing returns beyond a certain number of trees, with a trade-off in training efficiency as the computational cost per epoch increases. This suggests that an optimal balance exists between hierarchical complexity and computational feasibility for robust fine-tuning in hyperbolic space. Note that the multiple hierarchy trees do not affect the test-time inference efficiency.}

\begin{table}[h]
	\centering
	\renewcommand{\arraystretch}{1}
	\caption{Performance (\%) of different numbers of the hyperbolic trees with the average training time per epoch.}
	\vspace{-0.3cm}
	\resizebox{0.5\linewidth}{!}{
		\begin{tabular}{ccccc}
			\toprule
			\makecell{Number of Hyperbolic Trees} & Clean & PGD & AA & Time (min)  \\
			\midrule
			\rowcolor{LightBlue}1 & 62.14 & 44.34 & 42.72 & 73.2 \\
			3 & 62.36 & 44.78 & 42.97 & 88.0 \\
			\rowcolor{LightBlue}5 & 62.49 & 45.39 & 43.18 & 97.4 \\
			10 & 62.38 & 45.52 & 43.14 & 127.5 \\
			
			\bottomrule
		\end{tabular}
	}
	\label{tab:multi_trees}
\end{table}

\section{Different Weighting Mechanisms}
\label{app:weighting}
We explore the efficacy of using re-weighting for hierarchical image-text alignment in Eq. \eqref{eq:13}, comparing equal weighting and linear weighting ($\omega_l = 1-\frac{l}{L+1}$). Table \ref{tab:14} shows that emphasizing lower-level hierarchies (more fine-grained information) enhances zero-shot adversarial robustness. 
Lower (finer) level provides harder negatives for stronger semantic separation (crucial in improving robustness). We also include using only top 1-3  i) fine-grained levels, ii) superclass levels.

\begin{table}[h]
	\centering
	\renewcommand{\arraystretch}{1}
	\caption{Performance (\%) of different weighting mechanisms.}
	\vspace{-0.3cm}
	\resizebox{0.5\linewidth}{!}{
		\begin{tabular}{cccc}
			\toprule
			\makecell{Hierarchical Weighting Strategy} & Clean & PGD & AA  \\
			\midrule
Equal Weighting & 62.08 & 43.59 & 41.96 \\
Top 1-3 superclass levels & 61.82 & 43.71 & 41.84 \\
Top 1-3 fine-grained levels & 62.05 & 43.87 & 42.11 \\
\rowcolor{LightBlue}Linear Weighting & \textbf{62.14} & \textbf{44.34} & \textbf{42.72} \\
			
			\bottomrule
		\end{tabular}
	}
	\label{tab:14}
\end{table}

\section{Different Strategies to Represent Base Classes.} 
\label{app:averaging}
We explore three strategies for representing base-class embeddings during fine-tuning: (i) selecting a random instance from the base class, (ii) averaging in the Euclidean space followed by projection into the hyperbolic space, and (iii) hyperbolic averaging. Table \ref{tab:11} shows that aggregating via hyperbolic averaging produces the best zero-shot performance. 

\begin{table}[h]
	\centering
	\renewcommand{\arraystretch}{1}
	\caption{Performance (\%) of diverse base class representations.}
	\vspace{-0.3cm}
	\resizebox{0.45\linewidth}{!}{
		\begin{tabular}{cccc}
			\toprule
			\makecell{Base Class Representation} & Clean & PGD & AA  \\
			\midrule
			Random Instance & 58.92 & 41.65 & 40.12 \\
			Euclidean Averaging & 61.48 & 43.72 & 42.25 \\
			\rowcolor{LightBlue}Hyperbolic Averaging & \textbf{62.14} & \textbf{44.34} & \textbf{42.72} \\
			
			\bottomrule
		\end{tabular}
	}
	\label{tab:11}
\end{table}

\section{Robustness Against Black-box Adversarial Attacks}
\label{app:black-box}
In addition to zero-shot robustness evaluations against strong white-box attacks, we also analyze the black-box robustness of diverse adversarial fine-tuning methods using two practical multi-modal adversarial attacks: AFS \citelatex{supp_inkawhich2023adversarial} and MF-it \citelatex{supp_zhao2023evaluating}, as presented in Table \ref{tab:Blackbox_Adv}. Our method consistently achieves the best black-box robustness.

\begin{table}[h]
	\centering
	\renewcommand{\arraystretch}{1}
	\caption{\jh{Performance (\%) on black-box multimodal adversaries.}}
	\vspace{-0.3cm}
	\resizebox{0.55\linewidth}{!}{
		\begin{tabular}{ccccc}
			\toprule
			Evaluation Method & TeCoA \citelatex{supp_MaoGYWV23} & PMG-FT \citelatex{supp_wang2024pre} & FARE \citelatex{supp_schlarmann2024robust} & \cellcolor{LightBlue}\textbf{Ours} \\
			\midrule
			AFS \citelatex{supp_inkawhich2023adversarial} & 52.09 & 53.62 & 53.10 & \cellcolor{LightBlue}\textbf{57.92} \\
			MF-it \citelatex{supp_zhao2023evaluating} & 51.76 & 53.25 & 53.47 & \cellcolor{LightBlue}\textbf{57.38} \\
			\bottomrule
		\end{tabular}
	}
	\label{tab:Blackbox_Adv}
\end{table}

\section{The need for Hyperbolic space}

\begin{table}[h]
\caption{Comparisons of the Hyperbolic and the Euclidean variants, including naive hierarchical models.}
\label{tab:reb1}
\centering
\vspace{-0.3cm}
\renewcommand{\arraystretch}{0.5}
\resizebox{1\linewidth}{!}{
\begin{tabular}{ccccccc}
\toprule
\multirow{2}{*}{Type} & \multicolumn{3}{c}{Base Class} & \multicolumn{3}{c}{Superclass} \\
\cmidrule(lr){2-4} \cmidrule(lr){5-7}
& Clean & Robust & \textit{Transfer} & Clean & Robust & \textit{Transfer} \\
\midrule
Baseline (TeCoA) & 52.62 & 37.62 & 46.58 & 61.80 & 47.20 & 55.27 \\
\midrule
Hierarchical Euclidean SoftMax & 53.24 & 38.16 & 48.09 & 68.31 & 55.38 & 63.94 \\
Hierarchical Euclidean SoftMax +  $\omega_l$ (weighting) & 55.35 & 38.79 & 48.52 & 68.63 & 55.79 & 64.17 \\
Hier. Euclid. SoftMax + $\omega_l$ + Eq. \eqref{eq:14}-\eqref{eq:9} & 56.08 & 38.92 & 50.11 & 68.29 & 55.87 & 64.43 \\
%
%
Hier. Euclid. SoftMax + $\omega_l$ + Eq. \eqref{eq:14}-\eqref{eq:9} +Temp. $\tau$ & 55.93 & 39.13 & 50.73 & 67.91 & 55.08 & 64.56 \\
Hier. Euclid. SoftMax + $\omega_l$ + Eq. \eqref{eq:14}-\eqref{eq:9} +Temp. $2^l\tau$ & 57.17 & 39.45 & 51.05 & 68.46 & 55.85 & 64.73 \\
Hier. Euclid. SoftMax + $\omega_l$ + Eq. \eqref{eq:14}-\eqref{eq:9} +Temp. $(\tau_1,\ldots,\tau_L)$ & 57.67 & 39.90 & 52.23 & 68.75 & 56.04 & 64.97 \\
\textbf{Hyperbolic Space} ($\xi=1\times 10^{-4}$) & 61.70 & 42.60 & 62.43 & 71.15 & 56.71 & 65.82 \\
\textbf{Hyperbolic Space} ($\xi=1\times 10^{-6}$) & 61.24 & 42.05 & 60.59 & 70.44 & 55.93 & 63.26 \\
\rowcolor{LightBlue}{\textbf{Hyperbolic Space (Ours)} ($\xi=1\times 10^{-5}$)} & \textbf{62.14} & \textbf{42.72} & \textbf{63.34} & \textbf{71.68} & \textbf{57.13} & \textbf{66.40} \\
\bottomrule
\end{tabular}
}
\end{table}

Our classification margin induced by the Hyperbolic geometry differs from the Euclidean margin. Theorem \ref{th:margins} shows 
our margin range grows rapidly with the feature norm (unbounded at the Poincaré ball boundary) but the Euclidean SoftMax is bounded. The hierarchical level of feature vector is proportional to its norm, thus our design forms several generalization levels per sample \textbf{producing hierarchically-robust immunizing adversaries}. 
Table \ref{tab:reb1} includes \textbf{multi-level hierarchy-aware} (Eq. \eqref{eq:13}-\eqref{eq:9})  
Euclidean SoftMax which performs  worse.$\!\!\!$ 

To adjust margin, the Hierarchical Euclidean SoftMax needs to emply tuned SoftMax Temperature $\tau$ (row 5), or mimicking margin ranges from  Fig. \ref{fig:hyp_margin} (Temp. $2^l\tau$ in row 6). Tuning individual Temp. $\tau_1,\ldots,\tau_L$ for $L=5$ (row 7) takes $4^5=1024$ jobs. 
Our Hyperbolic model enjoys smarter margins as shown in Fig. \ref{fig:hyp_margin}. Note also the Poincaré stability \wrt $\xi$.

\section{AutoAttack Evaluations}
Table \ref{tab:reb2} provides AutoAttack-based comparisons.

\begin{table}[h]
\caption{AutoAttack results.}
\label{tab:reb2}
\centering
\vspace{-0.3cm}
\renewcommand{\arraystretch}{0.5}
\resizebox{1\linewidth}{!}{
\begin{tabular}{cccccccccccccccc>{\columncolor{LightBlue}}c}
\toprule
\textbf{Method} & \rotatebox{90}{ImageNet} & \rotatebox{90}{STL10} & \rotatebox{90}{CIFAR10} & \rotatebox{90}{CIFAR100} & \rotatebox{90}{SUN397} & \rotatebox{90}{Cars} & \rotatebox{90}{Food101} & \rotatebox{90}{OxfordPet} & \rotatebox{90}{Flower102} & \rotatebox{90}{DTD} & \rotatebox{90}{EuroSat} & \rotatebox{90}{FGVC} & \rotatebox{90}{PCAM} & \rotatebox{90}{Caltech101} & \rotatebox{90}{Caltech256} & \textbf{Average} \\
\midrule
CLIP & 0.00 & 16.23 & 4.09 & 0.00 & 0.00 & 0.00 & 2.24 & 0.00 & 0.00 & 0.00 & 0.00 & 0.00 & 0.00 & 11.72 & 6.10 & 2.69 \\ 
\midrule
TeCoA  & 39.07 & 82.11 & 58.21 & 32.64 & 30.41 & 12.19 & 26.28 & 61.68 & 27.85 & 21.58 & 15.67 & 4.95 & 25.77 & 67.75 & 58.14 & 37.62 \\
PMG-FT & 35.89 & 82.56 & 59.89 & 33.55 & 29.68 & 15.47 & 29.46 & 61.79 & 30.47 & 21.71 & 13.76 & 5.11 & 24.92 & 69.11 & 57.99 & 38.09 \\
FARE & 28.58 & 83.76 & 63.73 & 37.72 & 25.07 & 16.90 & 31.29 & 56.21 & 29.52 & 23.16 &  9.56 & 3.85 & 21.93 & 68.54 & 57.87 & 37.18 \\
AoS & 44.15 & 84.53 & 65.32 & 37.98 & 31.01 & 20.13 & 32.56 & 66.21 & 34.39 & 24.36 & 16.06 & 7.05 & 35.21 & 71.58 & 62.16 & 42.18 \\
\rowcolor{LightBlue}\textbf{Ours} & 44.59 & 85.55 & 66.03 & 39.48 & 32.38 & 19.79 & 33.17 & 66.57 & 35.38 & 24.08 & 16.46 & 7.85 & 35.23 & 71.96 & 62.28 & 42.72 \\
\rowcolor{LightBlue}\textbf{Ours (5 trees)} & \textbf{46.19} & \textbf{86.67} & \textbf{67.55} & \textbf{40.59} & \textbf{32.52} & \textbf{21.16} & \textbf{34.47} & \textbf{66.57} & \textbf{35.99} & \textbf{25.43} & \textbf{17.66} & \textbf{9.02} & \textbf{36.84} & \textbf{73.24} & \textbf{63.85} & \textbf{43.85} \\
\bottomrule

\end{tabular}
}
\end{table}

\section{Batch size \vs cost}

Table \ref{reb:tab3} compares training our model on batch size i)  128 (one GPU) and ii) 512. 

\begin{table}[h]
	\centering
	\caption{Batch size and cost evaluations on ViT-B.}
	\vspace{-0.3cm}
\label{reb:tab3}
\renewcommand{\arraystretch}{0.5}
\resizebox{0.5\linewidth}{!}{
\begin{tabular}{ccccc}
\toprule
Batch Size & Method & Clean & AA & Time (min) \\
\midrule
\multirow{27}{*}{128} & TeCoA & 52.08 & 37.19 & 210.6 \\
& PMG-FT & 56.83 & 37.74 & 261.3 \\
& FARE & 59.12 & 36.61 & 229.4 \\
& AoS & 61.28 & 41.55 & 515.8 \\
& \cellcolor{LightBlue}\textbf{Ours} (3 steps) & \cellcolor{LightBlue}61.81 & \cellcolor{LightBlue}42.30 & \cellcolor{LightBlue}283.7 \\
& \cellcolor{LightBlue}\textbf{Ours} (2 steps) & \cellcolor{LightBlue}61.67 & \cellcolor{LightBlue}42.19 & \cellcolor{LightBlue}209.4 \\
\midrule
\multirow{27}{*}{512} & TeCoA & 52.62 & 37.62 & 54.8 \\
& PMG-FT & 57.36 & 38.09 & 67.9 \\
& FARE & 59.67 & 37.18 & 59.5 \\
& AoS & 61.70 & 42.18 & 131.2 \\
& \cellcolor{LightBlue}\textbf{Ours} (3 steps) & \cellcolor{LightBlue}62.14 & \cellcolor{LightBlue}42.72 & \cellcolor{LightBlue}73.1 \\
& \cellcolor{LightBlue}\textbf{Ours} (2 steps) & \cellcolor{LightBlue}62.06 & \cellcolor{LightBlue}42.50 & \cellcolor{LightBlue}53.9 \\
\bottomrule
\end{tabular}
}
\end{table}

\section{Results Under Noisy Label Hierarchy.} While  \textbf{it is easy to inspect the label tree} to prevent errors (the label space is small), we investigate the effect of noise on results. On ImageNet-1K, we randomly corrupt superclass assignments of \textbf{10/50/100/200} base classes. Table \ref{reb:tab4} shows that the degradation under noise is moderate. If all assignments were random, the model would be acting similarly to non-hierarchical model.

\begin{table}[h]
	\centering
	\caption{The impact of label noise on results.}
	\vspace{-0.3cm}
\label{reb:tab4}
\renewcommand{\arraystretch}{0.5}
\resizebox{0.4\linewidth}{!}{
\begin{tabular}{cccc}
\toprule
\makecell{Hallucinated Classes} & Clean & PGD & AA  \\
\midrule
200 & 60.01 & 46.41 & 43.10 \\
100 & 60.61 & 46.83 & 43.72 \\
50 & 60.97 & 47.36 & 44.08 \\
10 & 61.03 & 47.67 & 44.41 \\
\rowcolor{LightBlue}0 (Standard Setup) & \textbf{61.19} & \textbf{47.81} & \textbf{44.59} \\
\bottomrule
\end{tabular}
}
\end{table}

\section{Sensitivity to prompt length.}  

Table \ref{reb:tab5}  reports results for several prompt lengths to show that the performance of VPT trend is stable. 

\begin{table}[h]
	\centering
	\caption{Performance \vs the prompt length.}
	\vspace{-0.3cm}
\label{reb:tab5}
\renewcommand{\arraystretch}{0.5}
\resizebox{0.4\linewidth}{!}{
\begin{tabular}{cccc}
\toprule
Prompt Length & Clean & PGD & AA \\
\midrule
50 & 54.19 & 34.57 & 31.74 \\
\rowcolor{LightBlue}\textbf{100 (Ours)} & 54.70 & 34.97 & 32.29 \\
200 & 54.94 & 35.23 & 32.62 \\
\bottomrule
\end{tabular}
}
\end{table}

\section{Sensitivity to the number of epochs.} 

Table \ref{reb:tab6} shows results for varied number of epochs and report performance, showing that our gains persist and that the chosen schedule is near the saturation point.

\begin{table}[h]
	\centering
	\caption{Performance \vs the epochs.}
	\vspace{-0.3cm}
\label{reb:tab6}
\renewcommand{\arraystretch}{0.5}
\resizebox{0.4\linewidth}{!}{
\begin{tabular}{cccc}
\toprule
Training Epochs & Clean & PGD & AA \\
\midrule
5 & 61.72 & 43.90 & 42.25 \\
\rowcolor{LightBlue}\textbf{10 (Ours)} & 62.14 & 44.34 & 42.72 \\
20 & 62.13 & 44.59 & 42.88 \\
\bottomrule
\end{tabular}
}
\end{table}

\section{\jh{Extended Related Works}}
\label{supp:Extend_RW}
\subsection{\jh{Hierarchical Feature Alignment}}
\jh{Hierarchical feature alignment has demonstrated its efficacy across various areas by aligning or fusing multi-level feature so that models can combine coarse semantic context with fine-grained details, leading to richer feature representations \citelatex{supp_hariharan2015hypercolumns, supp_lin2017feature}. An emerging perspective of hierarchical feature alignment is aligning representations with the hierarchical structure in data itself, typically using the hyperbolic space modeling \citelatex{supp_cannon1997hyperbolic}. Khrulkov \etal \citelatex{supp_khrulkov2020hyperbolic} introduced the first hyperbolic image embedding frameworks along these lines, adding hyperbolic layers to convolutional networks and demonstrating that hyperbolic embeddings often outperform Euclidean embeddings in representing visual features when the data has latent hierarchical relationships. Beyond this single vision modality, Desai \citelatex{supp_desai2023hyperbolic} introduced the hyperbolic representations in the CLIP model \citelatex{supp_radford2021learning} to capture the visual-semantic hierarchy inside images and texts. Unlike previous studies emphasizing hierarchical feature alignment alone, our work investigates zero-shot adversarial robustness within CLIP and its variants while further examining how these models can be applied across diverse scenarios. Central to our approach is the explicit construction of data hierarchies bridging textual and visual modalities, thereby reinforcing robust feature representations that enhance zero-shot adversarial performance.}

\subsection{Vision-Language Model Attacks}
\jh{With the rise of Vision-Language Models (VLMs), a variety of adversarial techniques have been introduced to embed subtle perturbations into both visual and textual inputs, thereby undermining VLM inference \citelatex{supp_zhang2022towards, supp_lu2023set}. Notably, Zhang \etal \citelatex{supp_zhang2022towards} laid the groundwork for studying adversarial scenarios rooted in cross-modal inconsistencies. While white-box attacks have been extensively tested, several recent works have shifted attention to black-box settings, highlighting more realistic adversarial risks in practical deployments \citelatex{supp_inkawhich2023adversarial, supp_zhao2023evaluating}. In this paper, we concentrate on establishing zero-shot adversarial robustness for VLMs to defend against such multimodal adversarial attacks.}

\subsection{Vision-Language Model Robustness}
\jh{To counteract the potential security threats brought by adversarial samples, adversarial fine-tuning \citelatex{supp_MaoGYWV23, supp_schlarmann2024robust} has been proposed, typically employing Parameter-Efficient Fine-Tuning (PEFT) techniques \citelatex{supp_jia2022visual, supp_zhou2022coop,supp_zhang2024moreextremegradientboost,supp_ni2024pace,supp_zhang2025crossspectra,supp_Zhu_2025_CVPR}. Mao \etal \citelatex{supp_MaoGYWV23} pioneered this direction by employing text-guided contrastive learning to conduct adversarial image-text embedding-level matching. Wang \etal \citelatex{supp_wang2024pre} tackled the generalization degradation via feature-level regularization. Schlarmann \etal \citelatex{supp_schlarmann2024robust} developed an unsupervised robust learning framework for downstream tasks. Li \etal \citelatex{supp_li2024one} for robustness in few-shot learning. However, existing methods perform pair-wise image-text alignment of image embedding with its base category embedding, overlooking the benefit of hierarchical decision boundaries from hierarchical labels. Thus, we reformulate traditional adversarial fine-tuning into a hierarchical alignment scheme across image and text modalities based on hyperbolic embedding, thereby mitigating the inherent fixation on base categories. Note that we focus primarily on zero-shot robustness achieved by fine-tuning VLMs.}

\begin{figure*}[!t]
	\centering
	\begin{subfigure}[t]{0.8\linewidth} 
		\centering
		\includegraphics[width=1\linewidth]{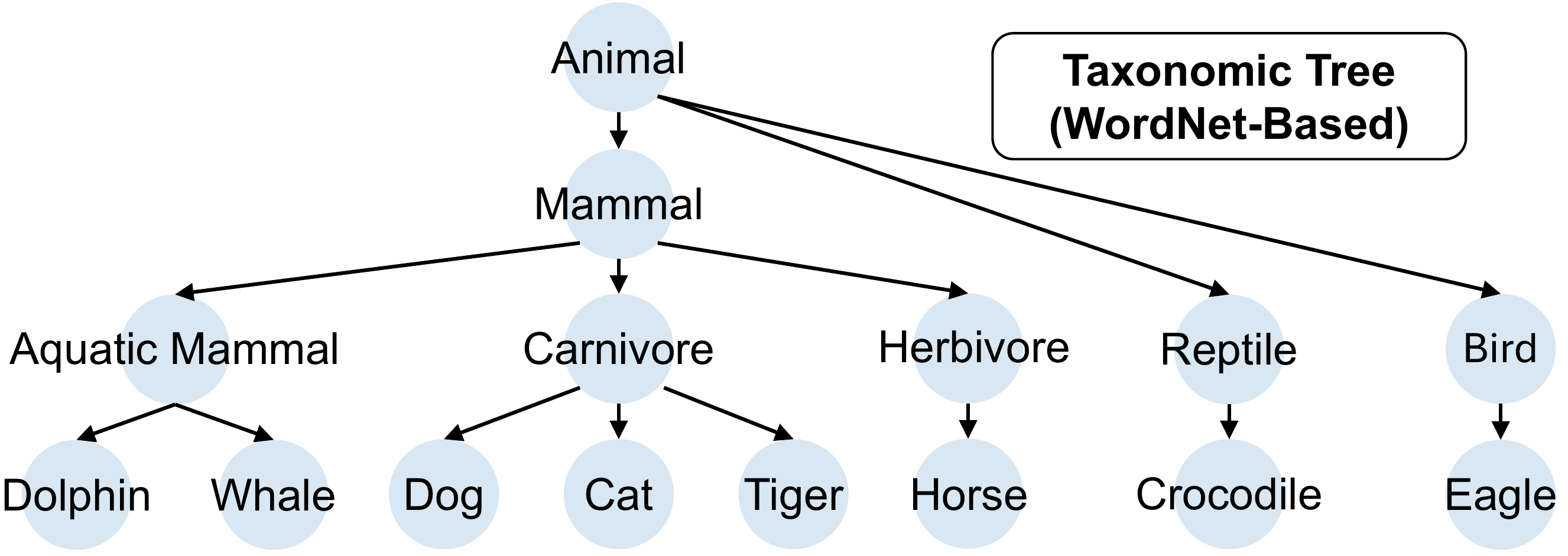}
		\caption{Taxonomic Tree (WordNet-Based)}
		\label{supp_fig:forest_1_WordNet}
	\end{subfigure}
	\begin{subfigure}[t]{0.8\linewidth} 
		\centering
		\includegraphics[width=1\linewidth]{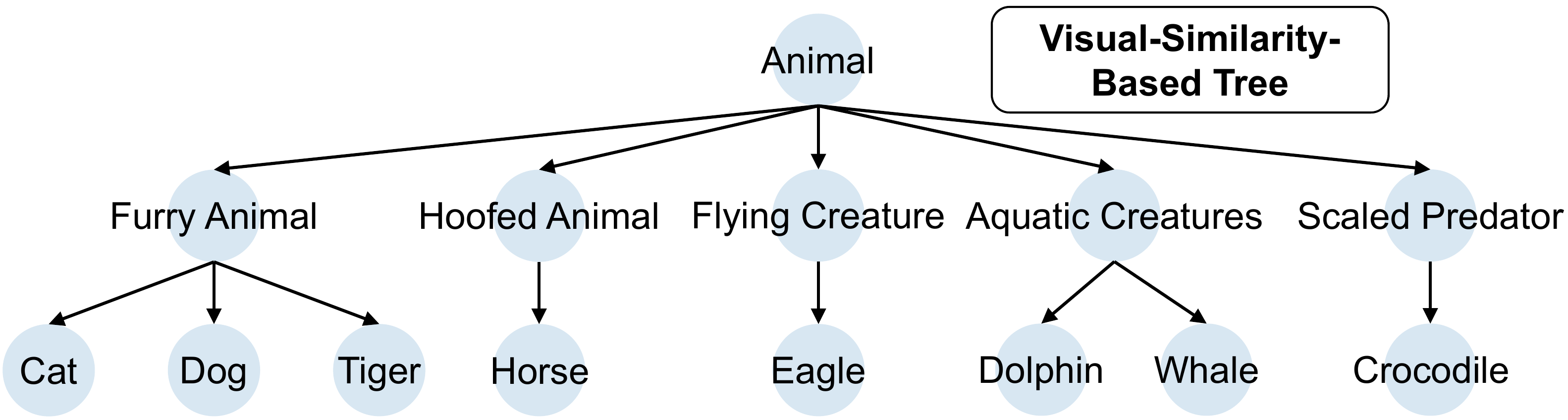}
		\caption{Visual-Similarity-Based Tree}
		\label{supp_fig:forest_2_Visual-Similarity}
	\end{subfigure}
	\begin{subfigure}[t]{0.8\linewidth} 
		\centering
		\includegraphics[width=1\linewidth]{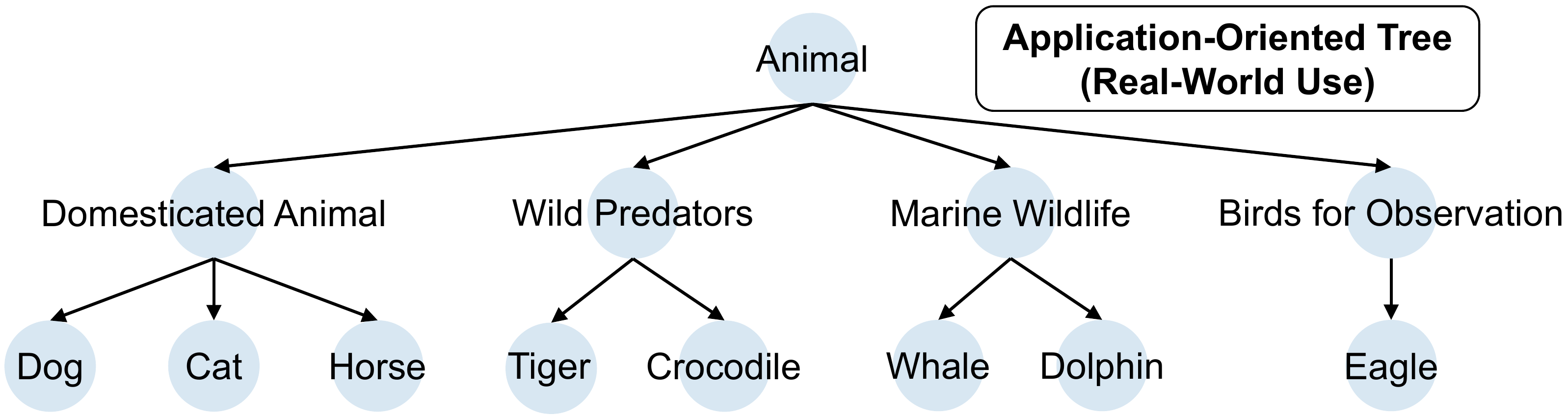}
		\caption{Application-Oriented Tree (Real-World Use)}
		\label{supp_fig:forest_3_Application}
	\end{subfigure}
	\caption{Diverse types of taxonomies \wrt base categories for different hierarchies.}
	\label{supp_fig:forests}
\end{figure*}

\section{\jh{Hierarchical Forests}}
\label{supp:Hierachical_forests}
Recall that we show that ensembling the hyperbolic (hierarchical) trees into hyperbolic forests improves adversarial robustness. To systematically explore the hierarchical structure of categories, we construct three distinct hierarchical forests based on different organizational principles, as illustrated in Figures \ref{supp_fig:forests}. Figure \ref{supp_fig:forest_1_WordNet} represents the taxonomic hierarchy, which follows the biological taxonomy derived from WordNet, grouping categories based on evolutionary relationships (\eg, mammals, reptiles, birds). Figure \ref{supp_fig:forest_2_Visual-Similarity} illustrates a visual similarity-based hierarchy, where categories are clustered according to shared physical attributes such as fur texture, body structure, or aquatic adaptation. This grouping better aligns with human visual perception and deep feature embeddings. Lastly, Figure \ref{supp_fig:forest_3_Application} depicts an application-oriented hierarchy, which organizes categories based on real-world interactions, such as domesticated animals (pets), wild predators, and aquatic species relevant to conservation and ecological studies. These three trees provide complementary perspectives for understanding hierarchical relationships, facilitating robust feature learning and adversarial robustness analysis in vision models.


{

	\small
	\bibliographystylelatex{ieeenat_fullname}
	\bibliographylatex{latex.bbl}

}

\end{document}